\documentclass[twoside,11pt]{article}

\usepackage{jmlr2e}
\usepackage{natbib}

\usepackage{enumitem}
\usepackage{times}
\usepackage{epsfig}
\usepackage{graphicx}
\usepackage{amsmath}
\usepackage{color}
\usepackage{amssymb}
\usepackage{adjustbox}
\usepackage{caption}
\usepackage{booktabs}
\graphicspath{{./}}

\usepackage{subcaption}
\usepackage{dirtytalk}
\usepackage[parfill]{parskip}
\usepackage{multirow}
\usepackage{float}

\newcommand{\ccc}[1]{\textcolor{black}{#1}}
\newcommand{\me}[1]{\textcolor{black}{#1}}

\newcommand{\omitme}[1]{}
\def\mypar#1{\vspace{1mm}{\bf #1.}\hspace{1mm}}

\usepackage{pgfplots,filecontents}
\pgfplotsset{compat=1.12}

\begin{document}

\title{Design Inspiration from Generative Networks}

\author{\begin{center}
    Othman Sbai$^{1,2,\footnote{Equal contributions}}$
\hspace{2.5mm}
Mohamed Elhoseiny$^{1,*}$
\hspace{2.5mm}
Antoine Bordes$^1$\\
Yann LeCun$^{1,3}$
\hspace{2.5mm}
Camille Couprie$^1$\\
$^1$ Facebook AI Research, $^2$ Ecole des ponts, Universit\'e Paris
Est, $^3$ New York University
\end{center}
}

\maketitle

\begin{abstract}
Can an algorithm create original and compelling fashion designs to serve as an inspirational assistant? To help answer this question, we design and investigate different image generation models associated with different loss functions to boost creativity in fashion generation.
  The dimensions of our explorations include: (i)  different Generative Adversarial Networks architectures that start from noise vectors to generate fashion items, (ii) novel loss functions that encourage creativity, inspired from Sharma-Mittal \me{divergence, a generalized mutual information measure for the widely used relative entropies} \ccc{ such as Kullback-Leibler}, and (iii) a generation process following the key elements of fashion design (disentangling shape and texture components).
A key challenge of this study is the evaluation of generated designs and the retrieval of best ones, hence we put together an evaluation protocol associating automatic metrics and human experimental studies that we hope will help ease future research.
 We show that our proposed creativity losses yield better overall appreciation than the one employed in Creative Adversarial Networks. In the end, about 61\% of our images are thought to be created by human designers rather than by a computer while also being considered original per our human subject experiments, and our proposed loss scores the highest compared to existing losses in both novelty and likability.
\end{abstract}

\begin{figure}[hb]
\begin{center}
\includegraphics[width=0.15\textwidth]{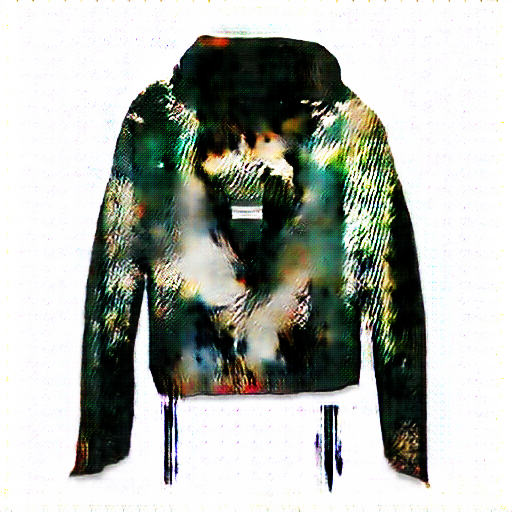}
\includegraphics[width=0.15\textwidth]{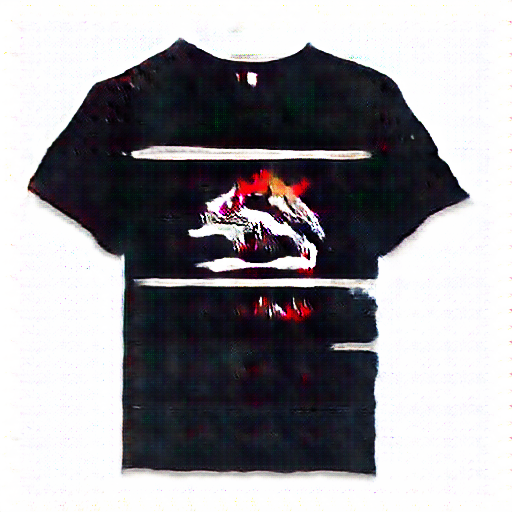}
\includegraphics[width=0.15\textwidth]{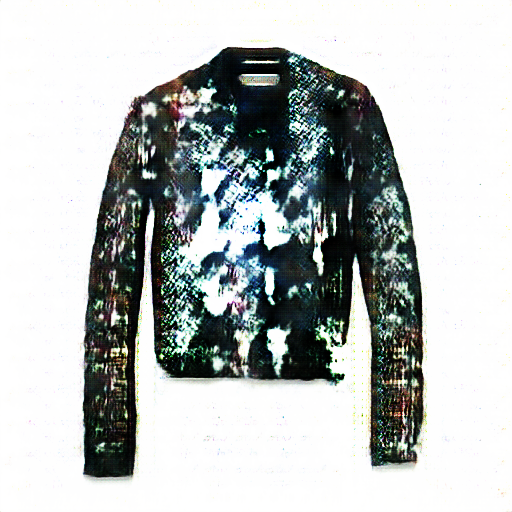}
\includegraphics[width=0.15\textwidth]{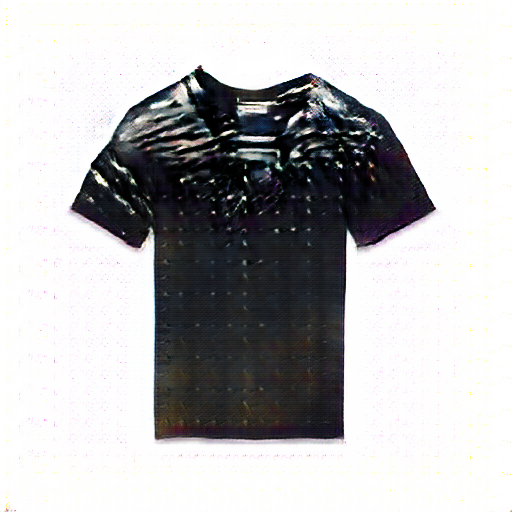}
\includegraphics[width=0.14\textwidth]{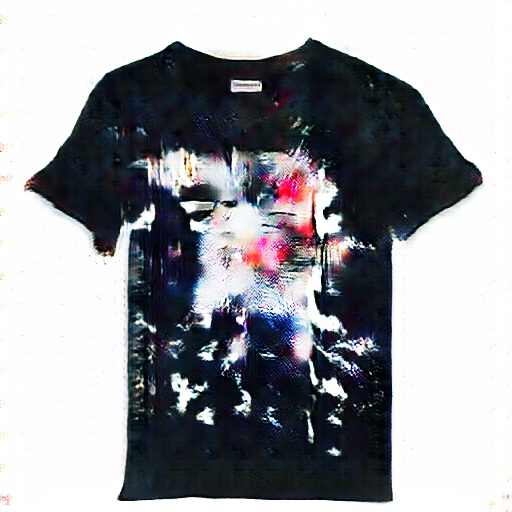}
\includegraphics[width=0.15\textwidth]{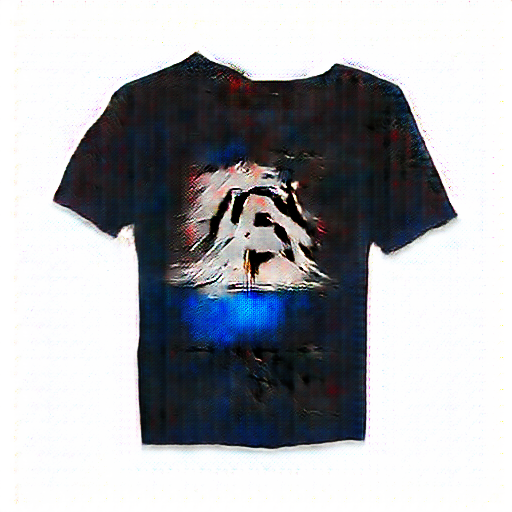}
\end{center}
\caption{Training generative adversarial models with appropriate losses leads to realistic and creative $512 \times 512$ fashion images.}
\label{fig:fig1_teaser}
\end{figure}


\section{Introduction}
\label{sec:introduction}

Artificial Intelligence (AI) research has been making huge progress in the machine's capability of human level understanding across the spectrum of perception, reasoning and planning~\citep{he2017mask,Andreas_2016_CVPR,silver2016mastering}.
Another key yet still relatively understudied direction is creativity where the goal is for machines to generate original items with realistic, aesthetic and/or thoughtful attributes, usually in artistic contexts.
We can indeed imagine AI to serve as inspiration for humans in the creative process and also to act as a sort of creative assistant able to help with more mundane tasks, especially in the digital domain.
Previous work has explored writing pop songs~\citep{briot2017deep}, imitating the styles of great painters~\citep{Gatys2016ImageStyleTransfer,dumoulin2016learned} or doodling sketches~\citep{ha2017neural} for instance.
However, it is not clear how {\it creative} such attempts can be considered since most of them mainly tend to mimic training samples without expressing much originality.
Creativity is a subjective notion that is hard to define and evaluate, and even harder for an artificial system to optimize for.
Colin Martindale put down a psychology based theory that explains human creativity in art~\citep{martindale1990clockwork} by connecting creativity or acceptability of an art piece to novelty with {\it ``the principle of least effort''}. As originality increases, people like the work more and more until it becomes too novel and too far from standards to be understood. When this happens, people do not find the work appealing anymore because a lack of understanding and of realism leads to a lack of appreciation.
This behavior can be illustrated by the Wundt curve that correlates the arousal potential (i.e. novelty) to hedonic responses (e.g. likability of the work) with an inverted U-shape curve.
Earlier works on computational creativity for generating paintings have been using genetic algorithms \citep{machadoiterative,machado2000nevar,mordvintsev2015inceptionism} to create new artworks by starting from existing human-generated ones and gradually altering them using pixel transformation functions.
There, the creativity is guided by pre-defined fitness functions that can be tuned to trade-off novelty and realism.
For instance, for generating portraits, \cite{dipaola2009incorporating} define a family of fitness functions based on specific painterly rules for portraits that can guide creation from emphasizing resemblance to existing paintings to promoting abstract forms.

Generative Adversarial Networks (GANs)~\citep{goodfellow2014generative,radford2015unsupervised} show a great capability to generate realistic images from scratch without requiring any existing sample to start the generation from.
They can be applied to generate artistic content, but their intrinsic creativity is limited because of their training process that encourages the generation of items close to the training data distribution; hence they show limited originality and overall creativity.
Similar conservative behavior can be seen in recent deep learning models for music generation where the systems are also mostly trained to reproduce pattern from training samples, like Bach chorales \citep{hadjeres2016deepbach}.
Creative Adversarial Networks (CANs, \cite{elgammal17}) have then been proposed to adapt GANs to generate creative content (paintings) by encouraging the model to deviate from existing painting styles.
Technically, CAN is a Deep Convolutional GAN (DCGAN) model~\citep{radford2015unsupervised} associated with an entropy loss that encourages novelty against known art styles.
The specific application domain of CANs allows for very abstract generations to be acceptable but, as a result, does reward originality a lot without judging much how such enhanced creativity can be mixed with realism and standards.

In this paper we study how AI can generate creative samples for fashion.
Fashion is an interesting domain because designing original garments requires a lot of creativity but with the constraints that items must be wearable.
For decades, fashion designers have been inventing styles and innovating on shapes and textures while respecting clothing standards (dimensions, etc.) making fashion a great playground for testing AI creative agents while also having the potential to impact everyday life.
In contrast to most generative models
works, the creativity angle we introduce makes us go
beyond replicating images seen during training.
Fashion image generation opens the door for breaking creativity into
design elements (shape and texture in our case), which is a
novel aspect of our work in contrast to CANs.
More specifically, this work explores various architectures and losses that encourage GANs to deviate from existing fashion styles covered in the training dataset, while still generating realistic pieces of clothing without needing any image as input.
In the fashion industry, the design process is traditionally organized around the collaboration between pattern makers, responsible for the material, the fabric and the texture,  and sample makers,  working on the shape models.
As detailed in Table~\ref{fig:exploration},  we follow a similar process in our exploration of losses and architectures.
We compare the relative impact of both shape and texture dimensions on final designs using
a comprehensive evaluation that jointly assesses novelty,  likeness and realism using both adapted automatic metrics,  and \ccc{extensive} humans subject experiments. 
To the best of our knowledge,  this work is the first attempt at incorporating creative fashion generation by explicitly relating it to its design elements. Our contributions are the following:

\begin{table}[t]
\begin{center}
{\small{\begin{tabular}{ccc}
\hline
Architecture & Creativity Loss & Design Elements\\
\hline
DCGAN~\citep{radford2015unsupervised} & CAN~\citep{elgammal17} &Texture (ours)\\
StackGAN (ours)  & MCE (ours) & Shape (ours)\\
StyleGAN (ours) & Bhattacharyya divergence (ours) & Shape \& Texture (ours) \\
\hline
\end{tabular}}}
\end{center}
\caption{Dimensions of our study. \ccc{We propose fashion image generation models that differ in their architecture and their creativity loss that encourages the generations to deviate from existing shapes,  textures,  or both.}}
\label{fig:exploration}

\end{table}

\begin{itemize}
\setlength\itemsep{-2.5mm}
\item We are the first to propose a creativity loss on image
generation of fashion items with a specific conditioning of
texture and shape, learning a deviation from existing ones. 
\item  We are the first to propose the {\it better leading
results, more general}, multi-class cross entropy criterion for learning to deviate
from existing shapes and textures. \ccc{We also express it as a particular occurrence of} the \me{Sharma-Mittal divergence as a generalized entropy that promotes generating creative fashion products}, \ccc{ allowing the exploration of various settings.}
\item We re-purpose automatic entropy based evaluation
criteria for assessment of fashion items in terms of
texture and shape; The correlations between the automatic
metrics that we proposed and our human study allow us
to {\it draw some conclusions with useful metrics revealing human
judgment}.
\item We propose a concrete solution to make our Style GAN
model \ccc{conditioned on shape images} work in a non-deterministic way, and trained it with
creative losses, resulting in a {\it novel and powerful model}.
\end{itemize}

As illustrated in Fig.~\ref{fig:fig1_teaser}, our best models manage to generate realistic images  with high resolution $512\times512$ using a relatively small dataset (about 4000 images).
More than 60\% of our generated designs are judged as being created by a human designer while also being considered original, showing that an AI could offer benefits serving as an efficient and inspirational assistant. A preliminary version of this work appears in \citep{Design} and is significantly extended here.


\section{Related work}

There has been a growing interest in generating images using convolutional
neural networks and adversarial training, given their ability to generate
appealing images unconditionally, or conditionally like from text, class labels,
and for paired and unpaired image
translations~\citep{Zhu2017cycleGANs}. GANs~\citep{goodfellow2014generative}
allow image generation from random numbers using two networks trained
simultaneously: a generator is trained to fool an adversarial network by
generating images of increasing realism. The initial resolution of generated
images was $32\times32$. From this seminal work, progresses were achieved in
generating higher resolution images, using a cascade of convolutional networks
\citep{denton2015laplacian} and deeper network architectures
\citep{radford2015unsupervised}. The introduced of auxiliary classifier GANs
\citep{Odena2017conditionalGAN} then consisted to add a label input in addition
to the noise and training the discriminator to classify the synthesized
$128\times128$ images. The addition of text inputs
\citep{Reeds2016whatwhere,Zhang2016StackGAN} allowed the generative network to
focus on the area of semantic interest and generate photo-realistic
$256\times256$ images. Recently, impressive $1024 \times 1024$ results were
obtained using a progressive growing of the generator and discriminator networks
\citep{karras2017progressive}, by training models during several tens of days.

Neural style transfer methods
\citep{Gatys2016ImageStyleTransfer,Johnson2016Perceptual} opened the door to the
application of existing styles on clothes \citep{Date2017Fashioning}, the
difference with generative models being the constraint to start from an existing
image in input.  \cite{Isola2016ImageToImage} relax this constraint partly by
starting from a binary image of edges, and present some generation of handbags
images. Another way to control the appearance of the result is to enforce some
similarity between input texture patch and the resulting image
\citep{Xian2017TextureGAN}. Using semantic segmentation and large datasets of
people wearing clothes, \citep{ZhuUrtasun2017iccvPrada,LassnerPG17} generate
full bodies images and are conditioning their outputs either on text
descriptions, color or pose information.  In this work, we are interested in
exploring the creativity of generative models and focus on presenting results
using only random or shape masks as inputs to leave freedom for a full
exploration of GANs creative power.

\section{Models: architectures and losses}
\label{sec:method}

Table~\ref{fig:exploration} summarizes our models and losses exploration.
Let us consider a dataset $\mathcal D$ of $N$ images.
Let $x_i$ be a real image sample and $z_i$ a vector of $n$ of real numbers sampled from a normal distribution. In practice $n=100$.

\subsection{GANs}

As in \citep{goodfellow2014generative,radford2015unsupervised}, the generator parameters $\theta_G$ are learned to compute examples classified as real by $D$:
\begin{equation*}
  \min_{\theta_G} \mathcal{L}_{G \mbox{ \scriptsize{real/fake}}}= \min_{\theta_G} \sum_{z_i\in \mathbb{R}^n} \log (1-D(G(z_i))).
\end{equation*}

The discriminator $D$, with parameters $\theta_D$, is trained to classify the true samples as 1 and the
generated ones as 0:

\begin{equation*}
  \min_{\theta_D} \mathcal{L}_{D\mbox{ \scriptsize{real/fake}}} = \min_{\theta_D} \sum_{\substack{x_i\in\mathcal{D},~z_i\in \mathbb{R}^n}} - \log D(x_i) - \log (1-D(G(z_i))).
  \label{Dr_loss}
\end{equation*}

\subsection{GANs with classification loss}

Following~\citep{Odena2017conditionalGAN}, we use  shape and texture labels to learn a shape classifier and a texture classifier in the discriminator. Adding these labels improves over the plain model and stabilizes the training for larger resolution.
Let us define the texture and shape integer labels of an image sample $x$ by
$\hat{t}$ and $\hat{s}$ respectively.
We are adding to the discriminator network either one branch for texture $D_t$ or shape $D_s$ classification or two branches for both shape and texture classification $D_{\{t,s\}}$. In the following section, for genericity, we employ the notation $D_{b,k}$, designating the output of the classification branch $b$ for class $k \in \{1,...,K\}$, where $K$ is the number of different possible classes of the considered branch (shape or texture).
We add to the discriminator loss the following classification loss:

\begin{equation*}
\mathcal{L}_{D} = \lambda_{D_r} \mathcal{L}_{D\mbox{ \scriptsize{real/fake}}} + \lambda_{D_b} \mathcal{L}_{D\mbox{\scriptsize{classif}}} \mbox{~ with~}
   \mathcal{L}_{D\mbox{ \scriptsize{classif}}} = - \sum_{x_i\in\mathcal {D}} \log \left(\frac{e^{D_{b,\hat{c}_i}(x_i)}}{\sum_{k=1}^K e^{D_{b,k}(x_i)}}\right),
\end{equation*}
where $\hat{c_i}$ is the label of the image $x_i$ for branch $b$.

\subsection{Creativity losses}

Because GANs learn to generate images very similar to the training images, we explore ways to make them deviate from this replication by studying the impact of an additional loss for the generator.  The final loss of the generator that is optimized jointly with $\mathcal{L}_{D}$ is:
\begin{equation*}
  \mathcal{L}_{G} = \lambda_{G_r} \mathcal{L}_{G\mbox{ \scriptsize{real/fake}}} + \lambda_{G_e} \mathcal{L}_{G\mbox{ \scriptsize{creativity}}}
\end{equation*}

\me{We explored different losses for creativity  that we detail in this section. } \ccc{First, we employ binary cross entropies over the adversarial network outputs as in CANs. Second, we suggest to employ the multi-class cross entropy (MCE) as a natural way to normalize the penalization across all classes. Finally, we show that this MCE loss is equivalent to a KL entropy and further generalize the creativity expression to the family of Sharma-Mittal divergences~\citep{IsSM07}.}

\omitme{This also includes a general  creativity loss based on Sharma–Mittal (SM) divergence~\citep{IsSM07}, a generalized mutual information measure for R\'enyi,
Tsallis, Bhattacharyya, and Kullback–Leibler (KL) relative entropies as we detail in this section. }


\mypar{Binary cross entropy loss (CAN \citep{elgammal17})}
Given the adversarial network's branch $D_c$ trained to classify different textures or shapes, we can use the CAN loss $\mathcal{L}_{\mbox{\scriptsize{CAN}}}$ as $\mathcal{L}_{G\mbox{ \scriptsize{creativity}}}$ to create a new style that confuses $D_b$:

\begin{equation}
\mathcal{L}_{\mbox{\scriptsize{CAN}}} = - \sum_{i} \sum_{k=1}^K \frac{1}{K} \log (\sigma (D_{b,k} (G(z_i))) )  + \frac{K-1}{K} \log(1-\sigma (D_{b,k}(G(z_i)))),
\end{equation}

where $\sigma$ denotes the sigmoid function.

\mypar{Multi-class Cross Entropy loss}
We propose to use as $\mathcal{L}_{G\mbox{ \scriptsize{creativity}}}$ the Multi-class Cross Entropy (MCE) loss between the class prediction of the discriminator and the uniform distribution. The goal is for the generator to make the generations hard to classify by the discriminator.
\begin{equation}
\mathcal{L}_{\mbox{\scriptsize{MCE}}} = -\sum_{i} \sum_{k=1}^K \frac{1}{K} \log \left(\frac{e^{D_{b,k}(G(z_i))}}{\sum_{q=1}^K e^{D_{b,q}(G(z_i))}}
\right) =-\sum_{i} \sum_{k=1}^K \frac{1}{K}\log (\hat{D_i}),
\label{eq_MCE}
\end{equation}

where \ccc{$\hat{D_i}$ is the softmax of $D_b(G(z_i))$}. 
In contrast to the CAN loss that treats every classification independently, the MCE loss should better exploit the class information in a global way. \omitme{ Both MCE and
sum of binary cross entropies losses encourage deviation from existing categories (shapes/textures here). Since the generator is
trained to output images that are not associated with
any label, the CAN loss can not contrast against negatives
and hence merely look at each dimension interdependently
encouraging it to get close to $\frac{1}{K}$.} \omitme{ ccc I removed the paragraph because: 1st sentense: redundant; 2nd "are not associated with any label" : true for MCE too. "can not contrast against negatives" : to rephrase: what are negative here?}

Our MCE loss is in fact equivalent to a Kullback-Leibler (KL) loss between a uniform distribution and the softmax output, since the entropy term of the uniform distribution is constant\me{; see derivation in Eq.~\ref{eq_KL}.}

\begin{equation}
 \me{ \mathcal{L}_{\mbox{\scriptsize{KL}}} = \sum_{i,k}  \frac{1}{K} \log (\frac{1}{K \hat{D_i}}) = \sum_{i,k}  \frac{1}{K} \log (\frac{1}{K})- \sum_{i,k} \frac{1}{K} \log (\hat{D_i}) 	\propto - \sum_{i,k} \frac{1}{K} \log (\hat{D_i}) = \mathcal{L}_{\mbox{\scriptsize{MCE}}}}
\label{eq_KL}
\end{equation}

It is more meaningful
conceptually to define the deviation over the joint distribution
over the categories which is what we modeled by
the KL divergences between the softmax probabilities and
the prior (defined as uniform in our case), as it also opens the door to study other
divergence measures. Finally, the MCE loss requires less compute comparing to CAN.

\begin{figure}[htb]
  \begin{center}
\includegraphics[width=0.16\linewidth]{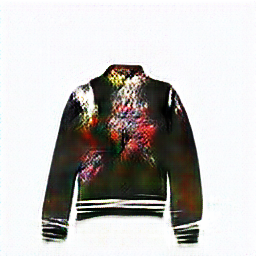}
\includegraphics[width=0.16\linewidth]{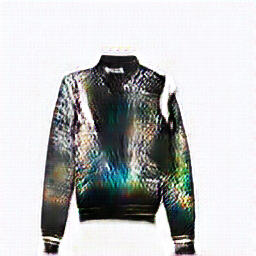}
\includegraphics[width=0.16\linewidth]{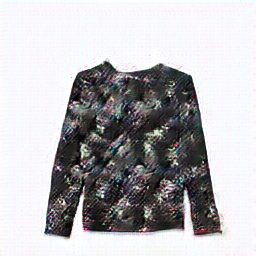}
\includegraphics[width=0.16\linewidth]{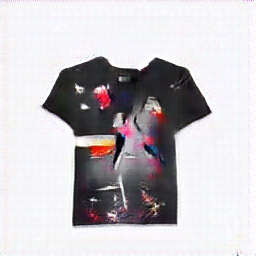}
\includegraphics[width=0.16\linewidth]{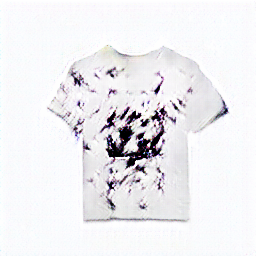}
\includegraphics[width=0.16\linewidth]{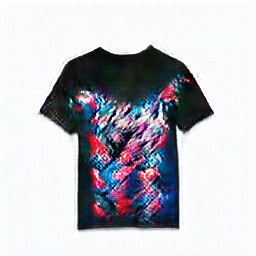}
  \end{center}
  \caption{Some generations from a DCGAN
    with MCE creativity loss applied on texture (Model GAN MCE tex).}
  \label{fig:best}
\end{figure}

\mypar{Generalized Sharma-Mittal Divergence}
We further generalize the expression of the creativity term to a broader family of divergences, unlocking new way of enforcing deviation from existing shapes and textures.
\cite{IsSM07} studied an entropy measure called Sharma-Mittal \omitme{on thermostatics in 2007,}which was originally introduced by \cite{SM75}. Given two parameters ($\alpha$ and $\beta$), 
 the Sharma-Mittal (SM) divergence $SM_{\alpha, \beta}(p\|q)$,  between two distributions $p$ and $q$ is defined $\forall \alpha >0,~\alpha \neq 1,~\beta \neq 1$ as 
\begin{equation}
SM(\alpha, \beta)(p || q) = \frac{1}{\beta-1} \left[ \sum_i (p_i^{1-\alpha} {q_i}^{\alpha})^\frac{1-\beta}{1-\alpha} -1\right]
\label{eq:smgen}
\end{equation}
It was shown in~~\citep{IsSM07} that most of the widely used divergence measures are special cases of SM divergence. For instance, each of the R\'enyi, Tsallis and Kullback-Leibler (KL) divergences can be defined as limiting cases of SM divergence as follows:
\begin{equation}
\small
\begin{split}
R_{\alpha}(p\|q) = &\lim_{\beta \to 1} {SM_{\alpha, \beta}(p\|q) }  = \frac{1}{\alpha-1} \ln (\sum_i {p_i^\alpha q_i^{1-\alpha} } )),  \\
T_{\alpha}(p\|q) = &\lim_{\beta \to \alpha} {SM_{\alpha, \beta}(p\|q) }= \frac{1}{\alpha-1} (\sum_i {p_i^\alpha q_i^{1-\alpha} } ) -1),  \\
KL(p\|q) = & \lim_{\beta \to 1, \alpha \to 1} {SM_{\alpha, \beta}(p\|q) } = \sum_i{p_i \ln ( \frac{p_i}{q_i}}).
\end{split}
\omitme{\begin{split}
R_{\alpha}(p\|q) = &\lim_{\beta \to 1} {SM_{\alpha, \beta}(p\|q) }  = \frac{1}{\alpha-1} \ln (\int_{-\infty}^{\infty}{p(t)^\alpha q(t)^{1-\alpha} dt} )),  \\
T_{\alpha}(p\|q) = &{SM_{\alpha, \alpha}(p\|q) }= \frac{1}{\alpha-1} (\int_{-\infty}^{\infty}{p(t)^\alpha q(t)^{1-\alpha} dt} ) -1),  \\
KL(p\|q) = & \lim_{\beta \to 1, \alpha \to 1} {SM_{\alpha, \beta}(p\|q) } = \int_{-\infty}^{\infty}{p(t) \ln ( \frac{p(t)}{q(t)} dt})
\end{split}}
\label{eqdef}
\end{equation}
\ccc{In particular, the } Bhattacharyya divergence ~\citep{bhatt67}, denoted by $B(p\|q)$   is a limit case of SM and R\'enyi divergences as follows as $\beta \to 1, \alpha \to 0.5$
\begin{equation*}
\small
\omitme{\begin{split}
B(p\|q)& = 2 \lim_{\beta \to 1, \alpha \to 0.5}  SM_{\alpha,\beta}(p\|q)  = - \ln \Big( \int_{-\infty}^{\infty} p(x)^{0.5} q(x)^{0.5} dx \Big).
\end{split}}
\begin{split}
B(p\|q)& = 2 \lim_{\beta \to 1, \alpha \to 0.5}  SM_{\alpha,\beta}(p\|q)  = - \ln \Big( \sum_i p_i^{0.5} q_i^{0.5} \Big).
\end{split}
\end{equation*}
Since the notion of creativity in our work is grounded to maximizing the deviation from existing shapes and textures through KL divergence in Eq~\ref{eq_KL}, we can  generalize our MCE creativity loss by minimizing  Sharma Mittal (SM) divergence   between a uniform distribution and the softmax output $\hat{D}$ as follows
\begin{equation}
\begin{split}
\mathcal{L}_{SM} = SM(\alpha,\beta)( \hat{D} || u) =
SM(\alpha, \beta)(\hat{D} || u) = \frac{1}{\beta-1} \sum_i (\frac{1}{K}^{1-\alpha} {\hat{D_i}}^{\alpha})^\frac{1-\beta}{1-\alpha} -1
\end{split}
\label{eq_sm}
\end{equation}
Note that the MCE loss in Eq~\ref{eq_KL} is a special case of the generalized SM loss where $\alpha \to 1$ and $\beta \to 1$ (i.e., KL divergence). Similarly, Tsallis , and Renyi, and Bhattacharyya losses are  special cases of SM loss where $\alpha = \beta$, $\beta \to 1$ and $\alpha \to 0.5$ and $\beta \to 1$. \me{We explored various parameters but we found both KL and  Bhattacharyya losses work the best and hence we focus on them in our experiments.} \ccc{We denote models trained with these losses by MCE and SM respectively.} 

\subsection{Network architectures}

We experiment using three architectures : modified versions of the DCGAN model~\citep{radford2015unsupervised},
StackGANs~\citep{Zhang2016StackGAN} with no text conditioning, and our proposed styleGAN.

\mypar{DCGAN} The DCGAN generator's architecture was only modified to output $256\times256$ or $512\times512$ images. The discriminator architecture also includes these modifications and contains additional classification branches depending on the employed loss function.

\mypar{Unconditional StackGAN}
Conditional StackGAN~\citep{Zhang2016StackGAN} has been proposed to generate  $256\times256$ images conditioned on captions\omitme{for mainly birds and flowers}. The method first generates a low resolution $64\times64$ image conditioned on text. Then, the generated image and the text are tiled on a $16\times16\times512$ feature map extracted from the $64\times64$ generated image to compute the final $256\times256$ image. We adapted the architecture by removing the conditional units (i.e. the text) but realized that it did not perform well for our application. The upsampling in ~\citep{Zhang2016StackGAN} was based on nearest neighbors which we found ineffective in our setting. Instead,  we first generate a low resolution image from normal noise using a DCGAN architecture~\citep{radford2015unsupervised}, then conditioning on it, we build a higher resolution image of $256\times256$ with a generator inspired from the pix2pix architecture with 4 residual blocks~\citep{Isola2016ImageToImage}. The upsampling was performed using transposed convolutions.
The details of the architecture are provided in the appendix.

\mypar{StyleGAN: Conditioning on masks}
To grant more control on the design of new items and get closer to standard fashion processes where shape and texture are handled by different specialists, we also introduce a model taking binary masks representing a desired shape in input.
Since that even for images on white background a simple threshold fails to extract accurate masks, we compute them using the graph based random walker algorithm \citep{Grady2006RW}.

\begin{figure}[htb]
\centering
\includegraphics[width=0.5\textwidth]{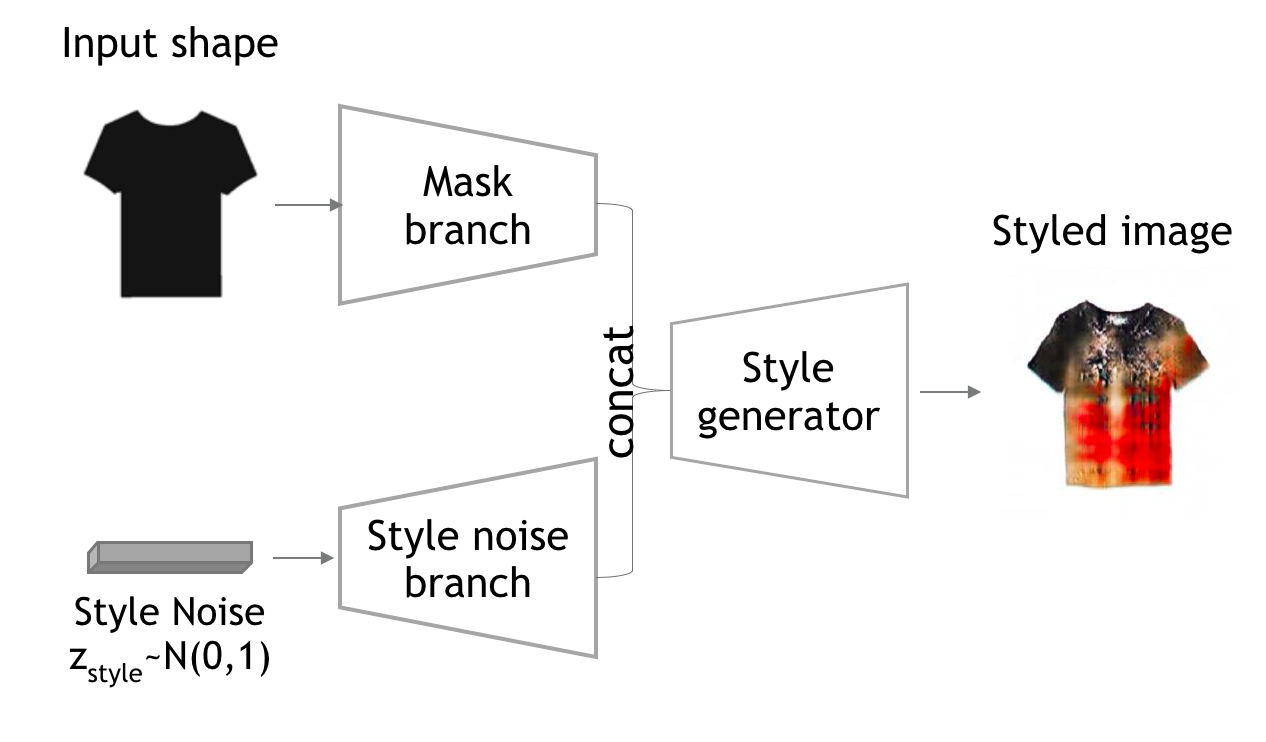}
\caption{From the segmented mask of a fashion item and different random vector $z$, our StyleGAN model generates different styled images.}
\label{fig:styleGANarchitecture}
\end{figure}

In the StyleGAN model, a generator is trained to compute realistic images from a
mask input and noise representing style information (see
Fig.~\ref{fig:styleGANarchitecture}), while a discriminator is trained to
differentiate real from fake images. We use the same discriminator architecture
as in DCGAN with classifier branches that learn shape and texture classification
on the real images on top of predicting real/fake discrimination.

Previous approaches of image to image translation such as
pix2pix~\citep{Isola2016ImageToImage} and CycleGAN~\citep{Zhu2017cycleGANs}
create a deterministic mapping between an input image to a single corresponding
one, i.e. edges to handbags for example or from one domain to another. This is
due to the difficulty of training a generator with two different inputs, namely
mask and style noise, and making sure that no input is being neglected.  In
order to allow sampling different textures for the same shape as a design need,
we avoid this deterministic mapping by enforcing an additional $\ell_1$ loss on
the generator:

\begin{equation}
\mathcal{L}_{rec} = \sum_i \sum_{p\in \mathcal{P}} |G(m_{i,p},z_i=0) - m_{i,p}|,
\end{equation}

where $m_{i,p}$ denotes the mask of a sample image $x_i$ at pixel $p$, and
$\mathcal{P}$ denotes the set of pixels of $m_i$. This loss \ccc{encourages the
  reconstruction of} the input mask in case of null input $z$ (i.e. zeros) and
hence ensures the disentanglement of the shape and
texture: having the reconstruction of the shape for null vector $z$ encourages the $z$
to only capture the texture variation. The architecture is detailed in the
appendix.


\begin{figure}[htb]
\centering
\includegraphics[height=0.265\textheight]{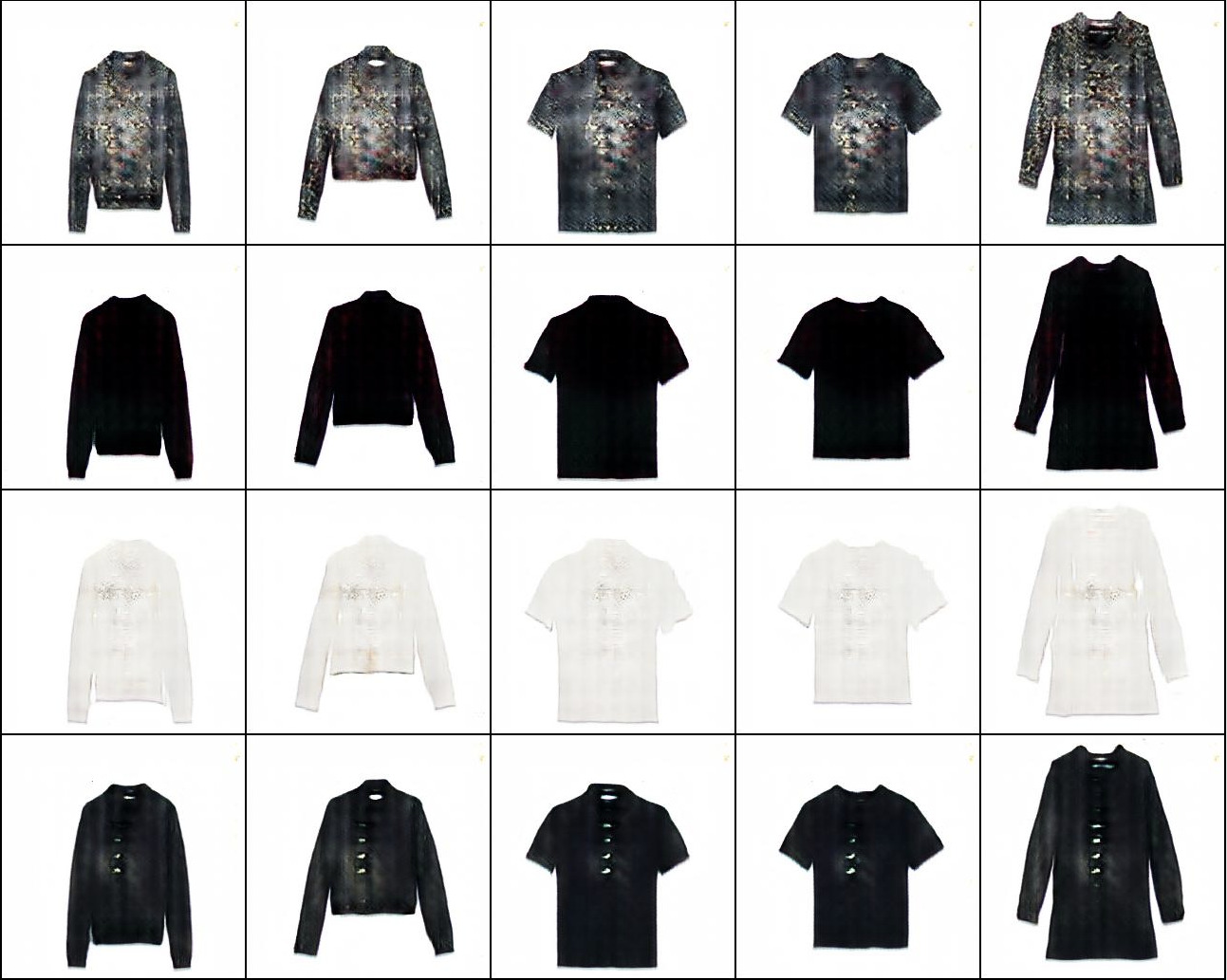} ~~ \includegraphics[height=0.265\textheight]{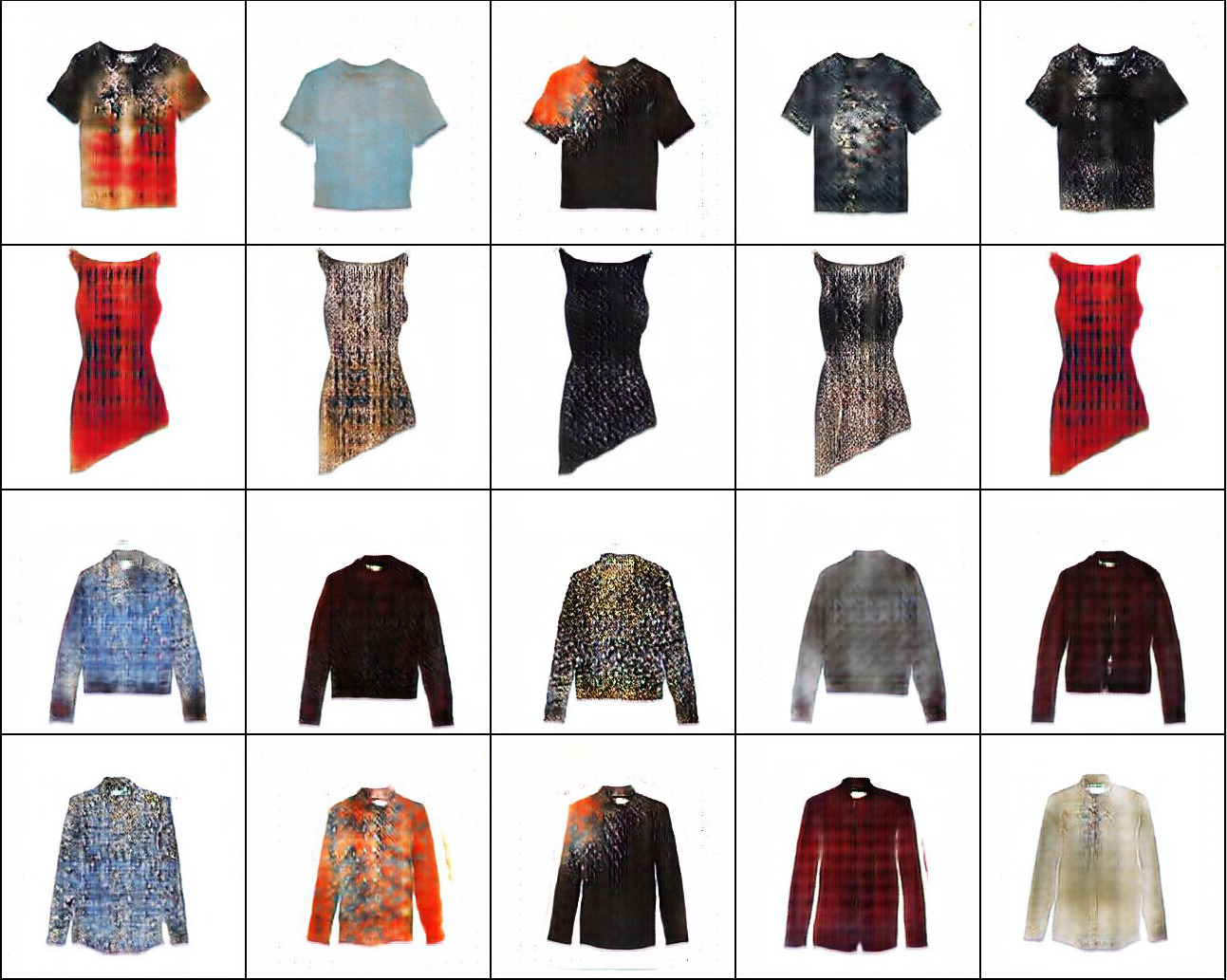}
\caption{From the mask of a product, our StyleGAN model generates different styled image for each style noise.}
\label{fig:styleGANdemo}
\end{figure}

\section{Experiments}
\label{sec:experiments}

\ccc{After presenting our datasets, we describe some automatic metrics we found useful to sort models in a first assessment, present quantitative results} followed by our human experiments that allow us to identify the best models.

\subsection{Datasets}
Unlike similar work focusing on fashion item generation~\citep{LassnerPG17,ZhuUrtasun2017iccvPrada}, we choose datasets which contain fashion items in uniform background allowing the trained models to learn features useful for creative generation without generating wearers face and the background.
We augment each dataset 5 times by jittering images with random scaling and translations.

\mypar{The RTW dataset}
We have at our disposal a set of 4157 images of different Ready To Wear (RTW) items of size
$3000\times3760$. Each piece is displayed on uniform white background. These images are classified into seven clothes categories: jackets, coats, shirts, tops, t-shirts, dresses and pullovers, and 7 texture categories: uniform, tiled, striped, animal skin, dotted, print and graphical patterns.

\mypar{Attribute discovery dataset}
We extracted from the Attribute discovery dataset~\citep{Berg2010AttributeBags} 5783 images of bags, keeping for our training only images with white background.
There are seven different handbags categories: shoulder, tote, clutch, backpack, satchel, wristlet, hobo. We also classify these images by texture into the same 7 texture classes as the RTW dataset.

\subsection{Automatic evaluation metrics}
\label{sec:eval}

Training generative models on fashion datasets generates impressive designs mixed with less impressive ones, requiring some effort of visual cherry-picking. We propose some automated criteria to evaluate trained models and compare different architectures and loss setups. We study in Section~\ref{sec:human} how these automatic metrics correlate with the human evaluation of generated images.

Evaluating the diversity and quality of a set of images has been tackled by scores such as the inception score and variants like the AM score~\citep{InceptionScoreTheoretical}. We adapt both of them for our two attributes specific to fashion design (shape and texture) and supplement them by a mean nearest neighbor distance. Our final set of automatic scores contains 10 metrics:

\begin{itemize}
\setlength\itemsep{-2mm}
\item Shape score and texture score, each based on a Resnet-18 classifier of (shape or texture respectively);
\item Shape AM score and texture AM score, based on the output of the same classifiers;
\item Distance to nearest neighbors images from the training set;
\item Texture and shape confusion of classifier;
\item Darkness, average intensity and skewness of images;
\end{itemize}


\mypar{Inception-like scores}
The Inception score \citep{salimans2016improved, warde2016improving} was introduced as a metric to evaluate the diversity and quality of generations, with respect to the output of a considered classifier~\citep{szegedy2017inception}. For evaluating $N$ samples $\{ x\}_{1}^{N}$, it is computed as
\begin{equation*}
I_{score}(\{ x\}_{1}^{N})= \exp(\mathop{\mathbb{E}}[KL(c(x) || \mathbb{E}[c(x)])])),
\end{equation*}

where $c(x)$ is the softmax output of the
trained classifier $c$, originally the Inception network.

Intuitively, the score increases with the confidence in the classifier prediction (low entropy score) of each image and with the diversity of all images (high overall classification entropy).
In this paper, we exploit the shape and texture class information from our two datasets to train two classifiers on top of Resnet-18 \citep{He2015Resnet18} features, leading to the \emph{shape score} and \emph{texture score}.

\mypar{AM scores}
We also use the AM score proposed in \citep{InceptionScoreTheoretical}. It improves over the inception score by taking into consideration the distribution of the training samples $\bar{x}$ as seen by the classifier $c$, which we denote $\bar{c}^{train} = \mathbb{E}[c(\bar{x})] $. The AM score is calculated as follows:
\begin{equation*}
\label{eq:AMScore}
AM_{score}(\{x\}_{1}^{N}) = \mathop{\mathbb{E}}[KL( \bar{c}^{train} || c(x)]
- KL(\bar{c}^{train} || \mathbb{E}[c(x)])]
\end{equation*}

The first term is maximized when each of the generated samples is far away from the overall training distribution, while the second term is minimized when the overall distribution of the generations is close to that of the training. In accordance with \citep{InceptionScoreTheoretical}, we find that this score is more sensible as it accounts for the training class distribution.

\mypar{Nearest neighbors distance}
To be able to assess the creativity of the different models while making sure
that they are not reproducing training samples, we compute the mean distance for each sample to its retrieved $k$-Nearest Neighbors (NN), with $k=10$, as the Euclidean distance between the features extracted from a Resnet18 pre-trained on ImageNet\citep{He2015Resnet18} by removing its last fully connected layer. These features are of size 512. This score gives an indicator of the similarity to the training data. A high NN distance may either mean that the generated images has some artifacts, in this case, it could be seen as an indicator of failure, or it could mean that the generation is novel and highly creative.

\begin{figure}[htb]
  \begin{center}
 \fbox{\includegraphics[width=0.45\linewidth]{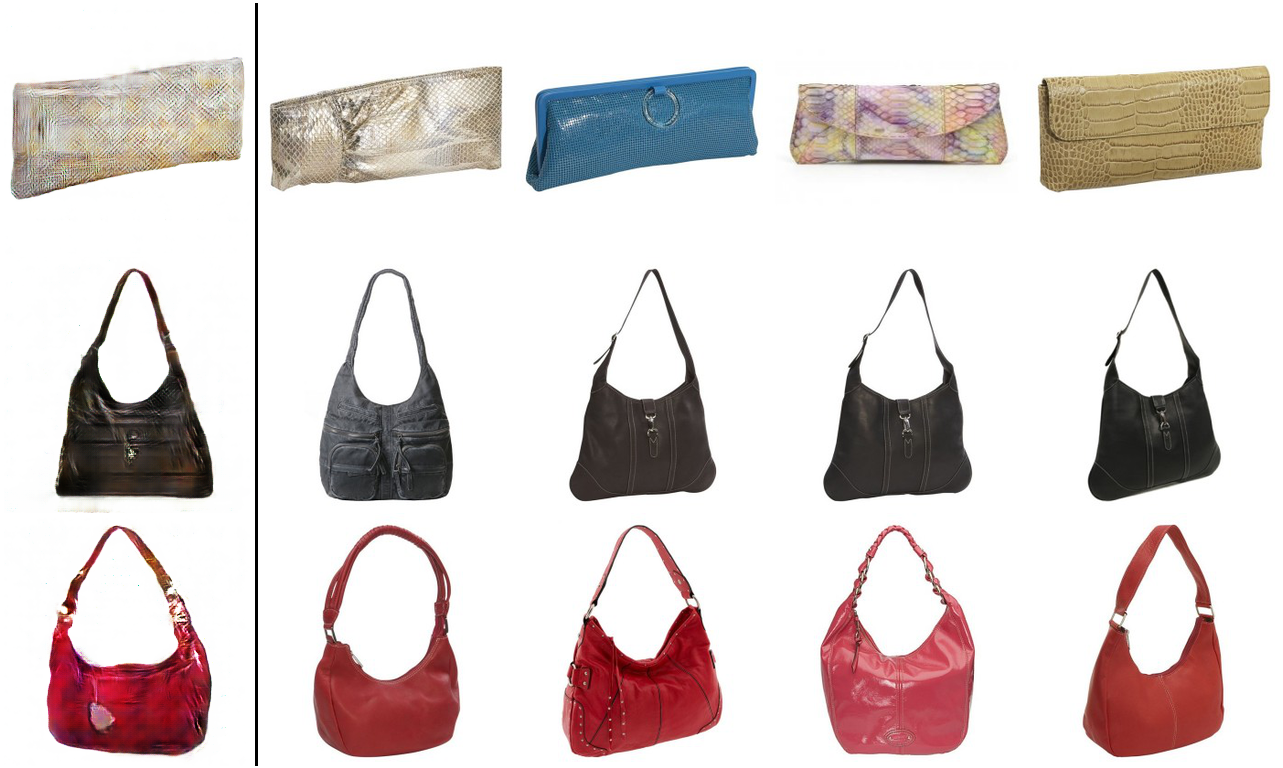}}
 \fbox{\includegraphics[width=0.45\linewidth]{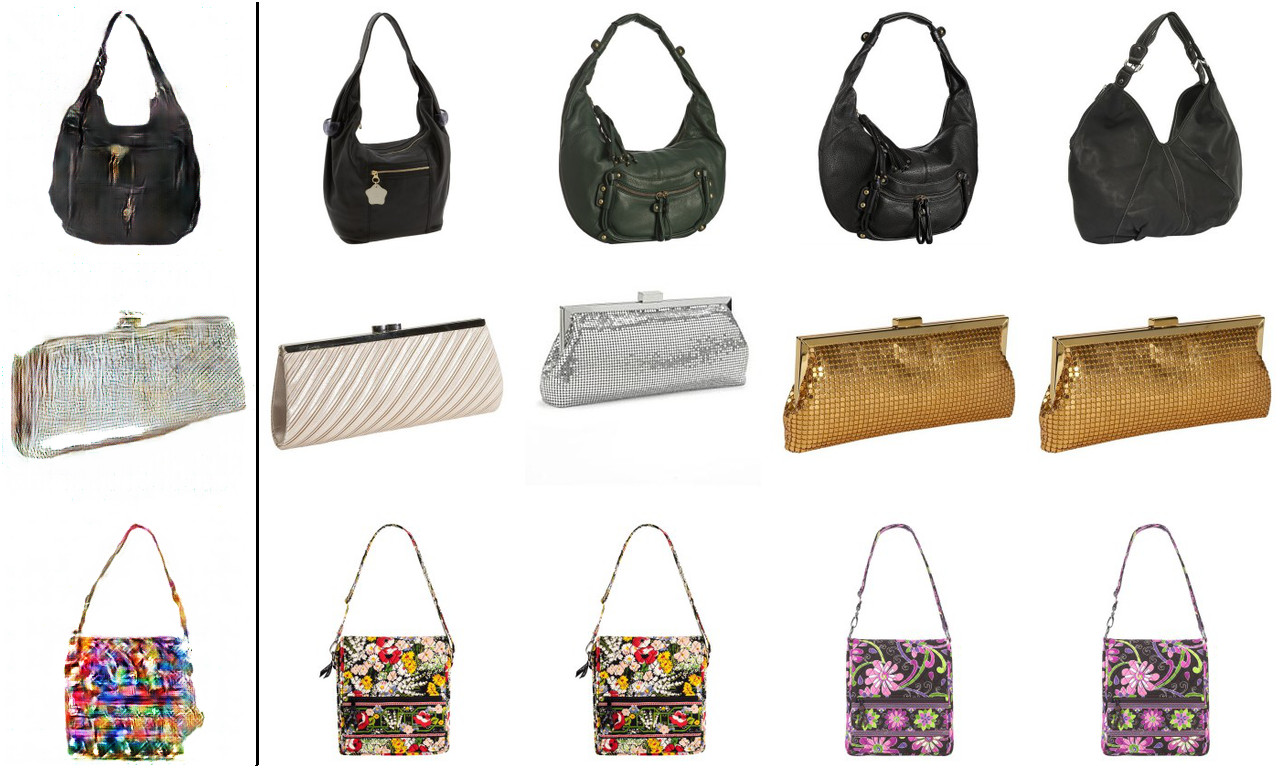}}
    \end{center}
    \caption{First column: random generations by the GAN MCE shape model. Four left columns: Retrieved Nearest Neighbors for each sample.}
    \label{fig:NN}
\end{figure}

\mypar{Darkness and Average intensity}
Dark colors might be preferred for fashion items as they are perceived as fitting well with other colors. From the $Y$ component of an image of a $YUV$ decomposition, this score counts the number of pixels whose brightness $Y$ is below a threshold (0.35). The average intensity score is also computed from the $Y$ component.

\mypar{Skewness}
The third order moment of the image histogram has been shown to be informative about image aesthetic~\citep{Motoyoshi2007,Attewell2007}. It computes the asymmetry of color intensity distribution. We chose to compute it from the histogram of the $Y$ channel of a $YUV$ decomposition.

\mypar{Shapes and texture confusion scores}
Given an image $x$, we may compute a score that reflects the confusion of the shape or texture classifier
\begin{equation}
C_{score}(x) = -\sum_{i=1}^K{\frac{e^{c_i(x)}}{\sum_k{e^{c_k(x)}}} \ln{\left( \frac{e^{c_i(x)}}{\sum_k{e^{c_k(x)}}}\right)}},
\end{equation}
where $c_k(x)$ denotes the output of classifier $c$. \ccc{This score is high when the texture or shape of the image is easy to identify.}

\mypar{Shapes and texture categories}

It might be useful to determine if the generation of particular categories of clothes (shapes or textures) plays a role into the likeliness of images. We therefore apply our shape and texture classifier to generated images.

\begin{table}[htb]
\begin{center}
\small{
\begin{tabular}{lccccccc}
\hline
& C sh & AM sh& C tex & AM tex &NN& Average& Dark \\
\hline
 GAN
   & 0.26 & 7.88   & 0.39 & 2.07 & \bf 14.2 & \bf 194  & 5719 \\
  GAN classif
   & 0.27 & \bf 9.78   & \bf 0.58 & 1.58 &13.0 & 188  & 7324 \\
 \hline
   GAN MCE sh
   & \bf 0.31 & 8.22  & \bf 0.59 & 1.69 & 13.1 & 192  & 6667 \\
  GAN MCE tex
   & 0.25 & 8.05   & \bf 0.59 & \bf 2.33 & \bf 13.8 & \bf 194   & 5617 \\
  GAN MCE shTex
   & 0.21 & 8.96   & 0.49 & 1.45 & 13.3 & 190 & 5843 \\ \hline
 CAN sh
   & 0.27 & 8.52   & 0.48 & 1.86 & 13.2 & 192  & 6524 \\
   CAN tex
   & 0.29 & 8.40   & 0.48 & \bf 2.24 & 13.4 & 190   & 6103 \\
  CAN shTex
   & 0.19 & \bf 10.1   & 0.46 & \bf 2.39 & 13.2 & 192   & 6958 \\ \hline
     SM tex
 &   0.23 & \bf 9.89  & 0.43 & \bf 2.44 & 12.9 & 193 & 7112 \\
   SM sh
   & 0.19 & \bf 9.39  & 0.51 & 1.21 & 12.3 & 191 & \bf 8663 \\
   SM shtex
    & 0.22 & 8.86  & 0.51 & 1.75 & \bf 14.1 & \bf 198 & 6738 \\
   \hline
  StackGAN
  & 0.25 & 8.82   & 0.52 & 1.95 & 12.9 & 193  & 7735 \\
  StackGAN MCE sh
   & 0.27 & 8.16  & \bf 0.64 & 2.03 & \bf 13.6 & \bf 195   & 6098 \\
  StackGAN MCE tex
   & 0.26 & 8.55   & \bf 0.63 & 1.68 & 13.2 & 186   & \bf 8483 \\
  StackGAN MCE shTex
   & 0.27 & 7.90   & \bf 0.71 & 1.91 & \bf 13.6 & \bf 194   & 4921 \\
  StackCAN sh
   & 0.20 & 8.96   & \bf 0.67 & 1.46 & 12.7 & 189   & \bf 9019 \\
  StackCAN tex
   & 0.22 & \bf 9.45   & 0.49 & \bf 2.32 & 13.2 & 189   & \bf 8796 \\
  StackCAN shTex
  & 0.26 & 8.11   & \bf 0.67 & \bf 2.07 & 13.4 & 193   & 5737 \\ \hline
  style GAN
   & 0.25 & 8.24   & 0.52 & 1.76 & \bf 13.7 & \bf 198   & 6622 \\
  style CAN tex
   & 0.29 & 7.93   & 0.48 & \bf 2.05 & \bf 13.9 & 193  & 6904 \\
  style GAN MCE tex
  & \bf 0.34 & 7.35   & 0.49 & 1.49 & 13.4 & 190   & \bf 9570 \\

 \hline
\end{tabular}\\
}
(a) Ready To Wear dataset\\
~\\
\small{
}

\begin{tabular}{lcccccccc}
\hline
& C sh & AM sh& C tex & AM tex &NN& Average& Dark \\
\hline
  GAN
   & \bf 0.43 & 3.65 & 1.71  & 1.57 & 20.8 & \bf 205   & \bf 2234 \\
  CAN tex
   & 0.35 & \bf 4.29 & 1.79   & 1.60 & \bf 21.4 & 201   & 1615 \\
  CAN sh
   & 0.39 & 4.23 & 1.88   & 1.26 & 21.0 & 201  &  \bf 2617 \\
  CAN sh tex
   & 0.39 & 3.93 & 1.89  & 1.56 & 21.1 & \bf 203  &  1954 \\
 GAN MCE tex
   & 0.34 & \bf 4.38 & \bf 1.99  & 1.60 & \bf 21.6 & 196   & \bf 2691 \\
  GAN MCE sh
  & 0.38 & 4.15 & \bf 1.98   & \bf 1.84 & 20.8 & \bf 207   & 1593 \\
  GAN MCE sh tex
  & \bf 0.42 & 3.73 & \bf 2.00   & \bf 1.80 & 21.0 & 200  & \bf 2333 \\
 \hline
\end{tabular}\\
(b) Attribute bags dataset
\end{center}
\caption{Quantitative evaluation on the RTW dataset and bag datasets.  \ccc{For better readability we only display metrics that correlate most with human judgment. Higher scores, highlighted in bold, are usually preferred.}}
\label{tab:automaticMetricsResults}
\end{table}

\subsection{Automatic evaluation results\omitme{quantitative}}

We experiment using weights $\lambda_{G_e}$ of 1 and 5 for the MCE creativity loss. It appeared that the weight 1 worked better for the bags dataset, and 5 for the RTW dataset. \omitme{We note, from the observation of NN distances in Table \ref{tab:automaticMetricsResults} that the RTW dataset has a much lower internal variability than the handbags dataset.} We also tried different weights for the CAN \ccc{and SM} loss  but they did not have a large influence on the results and was fixed to 1. All models were trained using the default learning rate $0.002$ \omitme{and an input noise size $n_z=100$} as in \citep{radford2015unsupervised}. Our different models take about half a day to train on 4 Nvidia P100 GPUs for $256\times256$ models and almost 2 days for the $512\times512$ ones. In our study, it was more convenient from a memory and computational resources standpoint to work with $256\times256$ images but we also provide $512\times512$ results in Fig.~\ref{fig:fig1_teaser} to demonstrate the capabilities of our approach. \ccc{For each setup, we manually select  four saved models after a sufficient number of iterations. Our models produce plausible results after training for 15000 iterations with a batch size of 64 images.}

Table \ref{tab:automaticMetricsResults} presents shape and texture classifier
confidence C scores, AM scores (for shape and texture), average NN distances
computed for each model, \ccc{Average intensity and Darkness scores on a set of
  100 randomly selected images}.  Our first observation is that the DCGAN model
alone seems to perform worse than all other tested models with the highest NN
distance and lower shape and texture scores. The value of the NN distance score
may have different meanings. A high value could mean an enhanced creativity of
the model, but also a higher failure rate.

For the RTW dataset, \ccc{the two models having high AM texture score, and high
  NN distances scores are DCGAN with MCE loss models and Style CAN with texture
  creativity.}  On the handbags datasets, the models obtaining the best metrics
overall are the DCGAN with \ccc{MCE creativity losses texture alone or on shape
  and texture.}  To show that our models are not reproducing exactly samples
from the dataset, we display in Fig.~\ref{fig:NN} results from the model having
the lowest NN distance score, with its 4 nearest neighbors. We note that even if
uniform bags tend to be similar to training data, complex prints show high
differences. These differences are amplified on the RTW dataset.

\mypar{Creating evaluation sets}

\ccc{Motivated by the will of accessing automatically the best generations from
  each model,} we extract different clusters of images with particular visual
properties that we want to associate with realism, overall appreciation and
creativity.  Given the selected models, 10000 images are generated from random
numbers -- or randomly selected masks for the styleGAN model -- to produce 8
sets of 100 images each.  Based on the shape entropy, texture entropy and mean
nearest neighbors distance of each image we can rank the generations and select
the ones with (i) high/low shape entropy, (ii) high/low texture entropy, (iii)
high/low NN distance to real images.  We also explore random and mixed sets such
as \emph{low shape entropy} and \emph{high nearest neighbors distance}. We
expect such a set to contain plausible generations since low shape entropy
usually correlates with well defined shapes, while high nearest neighbor
distance contains unusual designs. Overall, we have 8 different sets that may
overlap. We choose to evaluate 100 images for each set.

\subsection{Human evaluation results}
\label{sec:human}

\begin{figure}[htb]
\begin{center}
\includegraphics[width=0.95\linewidth]{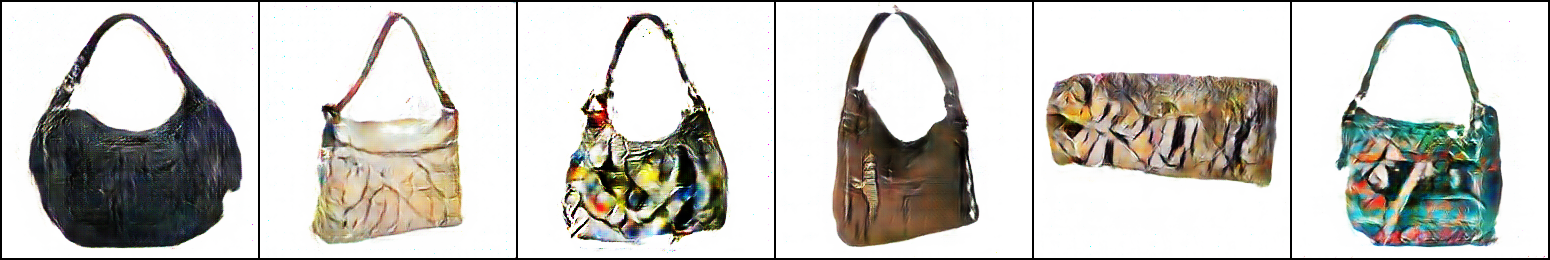}
\includegraphics[width=0.95\linewidth]{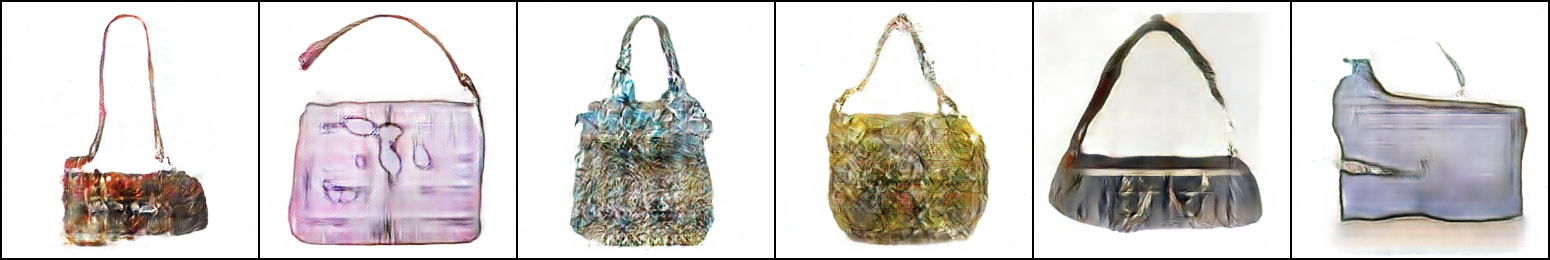}
\end{center}
\caption{Best generations of handbags as rated by human annotators. Each question is in a row. Q1: overall score, Q2: shape novelty.}
\label{fig:best_human_per_qst_handbags}
\end{figure}

We perform a human study where we evaluate different sets of generations of interest on a designed set of questions in order to explore the correlations with each of the proposed automatic metrics in choosing best models and ranking sound generations.
As our RTW garment dataset could not be made publicly available,
we conducted two independent studies:
\begin{enumerate}[leftmargin=*]
\item In our main human evaluation study, 800 images –- \ccc{selected as described in the previous section} –- per model
were evaluated, each image evaluated by 5 different
persons. There were in average 90 participants per model
assessment, resulting in average to 45 images assessed per
participant. Since the assessment was conducted given the
same conditions for all models, we are confident that the
comparative study is fair. Each subject is shown images from the 8 selected sets described in the previous section and is asked 6 questions:
\begin{itemize}
\setlength\itemsep{-2mm}
    \item Q1: how do you like this design overall on a  scale from 1 to 5?
    \item  Q2/Q3: rate the  novelty of shape (Q2) and texture (Q3) from 1 to 5.
    \item Q4/Q5: rate the complexity of shape (Q4) and texture (Q5) from 1 to 5.
    \item Q6: Do you think this image was created by a fashion designer or generated by computer? (yes/no)
\end{itemize}
Each image is annotated by taking the average rating of 5 annotators.
\item We conducted another study in our lab were we mixed
both generations (500 total images picked randomly from 5
best models) and 300 real down-sampled images from the RTW dataset. We asked if the images were real or generated to about 45 participants who rated 20 images each in average. We obtain 20\% of the generations thought to be real, and 21.5\% of the original dataset images were considered to be generated.
\end{enumerate}

\begin{figure}[htb]
\begin{tabular}{lc}
\begin{minipage}{.32\textwidth}
{\small Likeability}
\centering
  \begin{center}

\begin{tikzpicture}
{\small

    \begin{axis}[xlabel=Real appearance , height=8cm, width=6cm]
        \addplot+[
                visualization depends on={value \thisrow{nodes}\as\myvalue},
                scatter/classes={
                a={mark=text,text mark=\scriptsize{\myvalue},blue},
                b={mark=text,text mark=\scriptsize{\myvalue},red},
                c={mark=*,blue},
                d={mark=*,red}
                },
                scatter,draw=none,
                scatter src=explicit symbolic]
         table[x=x,y=y,meta=label]
            {data.txt};
    \end{axis}
    }
\end{tikzpicture}
  \end{center}
\end{minipage}
&
\begin{minipage}{.7\textwidth}
\centering
\begin{center}
\scriptsize{
\begin{tabular}{lcccccc}
\hline
Method/Human & over- & shape & shape & tex. & tex. & real\\
Method & all & nov. & comp. & nov. & comp. & fake\\
\hline
{DCGAN MCE shape}   & \textbf{3.78}   & \textbf{3.58}   & \textbf{3.57}  & \textbf{3.64}   & \textbf{3.57}   & \textbf{60.9}  \\ 
{DCGAN MCE tex}   & \textbf{3.72}   & \textbf{3.57}   & 3.52   & \textbf{3.61}   & \textbf{3.58}  & \textbf{61.1}  \\ 
{StyleCAN tex}   & 3.65   & 3.37   & 3.31   & 3.44   & 3.21   & 49.7  \\ 
{StackGAN}   & 3.62   & 3.45   & 3.38   & 3.43   & 3.33   & 51.9  \\ 
{StyleGAN MCE tex}   & 3.61   & 3.38   & 3.29   & 3.50   & 3.37   & 53.4 \\ 
{DCGAN SM tx}   & 3.60   &  3.43 & 3.42  &  3.46 & 3.41   & 52.6 \\ 
{StackGAN MCE tex}   & 3.59   & 3.36  & 3.31   & 3.44   & 3.28   & 55.9 \\ 
{StyleGAN}   & 3.59   & 3.28   & 3.21   & 3.27   & 3.15   & 47.2  \\ 
{StackGAN CAN sh}   & 3.51   & 3.56   & \textbf{3.56}   & 3.58   & 3.40   & 50.7  \\ 
{DCGAN MCE shTex}   & 3.49   & 3.40   & 3.24   & 3.40   & 3.31   & \textbf{61.3}  \\ 
{StackGAN CAN tex}   & 3.48   & \textbf{3.57}   & 3.54  & 3.55   & 3.50  & 48.4 \\ 
{DCGAN CAN shTex}   & 3.47   & 3.28   & 3.18   & 3.33   & 3.16   & \textbf{63.8}  \\ 
{StackGAN MCE sh}   & 3.45   & 3.27   & 3.16   & 3.28   & 3.12   & 60.4  \\ 
{StackGAN CAN shTex}   & 3.42   & 3.37   & 3.32   & 3.44   & 3.32   & 49.5  \\ 
{DCGAN classif}   & 3.42   & 3.32  & 3.32   & 3.37   & 3.29   & 52.7  \\ 
{ DCGAN SM sh}   & 3.39   &  3.27 & 3.12   &  3.30  & 3.23   & 55.1 \\ 
{DCGAN CAN tex}   & 3.37  & 3.23   & 3.12   & 3.35   & 3.09   & 59.7  \\ 
{DCGAN CAN sh}   & 3.33  & 3.28   & 3.16   & 3.27   & 3.12   & 55.0  \\ 
{StackGAN MCE shtex}   & 3.30   & 3.25   & 3.28   & 3.31   & 3.27   & 41.5  \\ 
{DCGAN}   & 3.22   & 2.95   & 2.78   & 3.24   & 2.83   & 60.4  \\
{DCGAN SM shTex}
   & 3.20   &  3.10 & 3.00   &  3.13  & 3.1   & 45.5 \\ 
\hline
\end{tabular}
}
\end{center}

\end{minipage}
\end{tabular}
\caption{Human evaluation on the 800 images from all sets ranked by decreasing overall score (higher is better) on the RTW dataset. Evaluation of the different models on the RTW dataset by human annotators on two axis: likability and real appearance. Our models using MCE, SM creativity or shape conditioning are highlighted by darker colors. They reach nice trade-offs between real appearance and likability.}
\label{tb:human_all_sets_overall}
\end{figure}

Table~\ref{tb:human_all_sets_overall} presents the average scores obtained by each model on each human evaluation question for the RTW dataset. From this table, we can see that using our creativity loss (MCE shape and MCE tex) performs better than the DCGAN baseline. We obtain a similar ranking of the models on the set of random images. The image set of 'low NN distance' contains the most popular images,  with an average of 4.16 overall score for the best model. We checked the statistical significance of the results of Table~\ref{tb:human_all_sets_overall} by computing paired Student t-tests. The obtained p-value between the overall scores of the two first ranked models is 0.03, and the one between the first and the third ranked models is lower than $10^{-6}$, allowing us to reject the hypothesis of statistically equal averages.

\omitme{While the two proposed models with MCE creativity loss rank the best in the RTW dataset on the overall score, we observe that the preferred images have low nearest neighbor distance  which may also be deduced from the correlation Table~\ref{tb:corrAll} showing Pearson correlation scores between the 5 automatic metrics and the 6 human evaluation questions. 
This means that generations which are not close to their nearest neighbors are not always pleasant. It is indeed a challenge to obtain models able to generate novel (high nearest neighbor distance) and at the same time pleasant generations. However, we observe that the models that score better in the high nearest neighbors distance set are clearly the ones with our creativity loss.}

Fig.~\ref{tb:human_all_sets_overall} shows how well our approaches worked on two axis: likability and real appearance. The most popular methods are obtained by the models employing creativity loss  and in particular our proposed MCE loss, as they are perceived as the most likely to be generated by designers, and the most liked overall.
We are greatly improving the state-of-the-art here, going from a score of 64\% to more than 75\% in likabity from classical GANs to our best model with shape creativity. Our proposed Style GAN and StackGAN models are producing competitive scores compared to the best DCGAN setups with high overall scores. In particular, our proposed StyleGAN model with creativity loss is ranked in the top-3, some results are presented in Fig.~\ref{fig:styleGANdemo}. We display images which obtained the best scores for each of the 6 questions in Fig.~\ref{fig:best_human_per_qst}, and some results of handbags generation in Fig.~\ref{fig:best_human_per_qst_handbags}.

\begin{figure}[htb]
  \begin{center}
\includegraphics[width=0.9\linewidth]{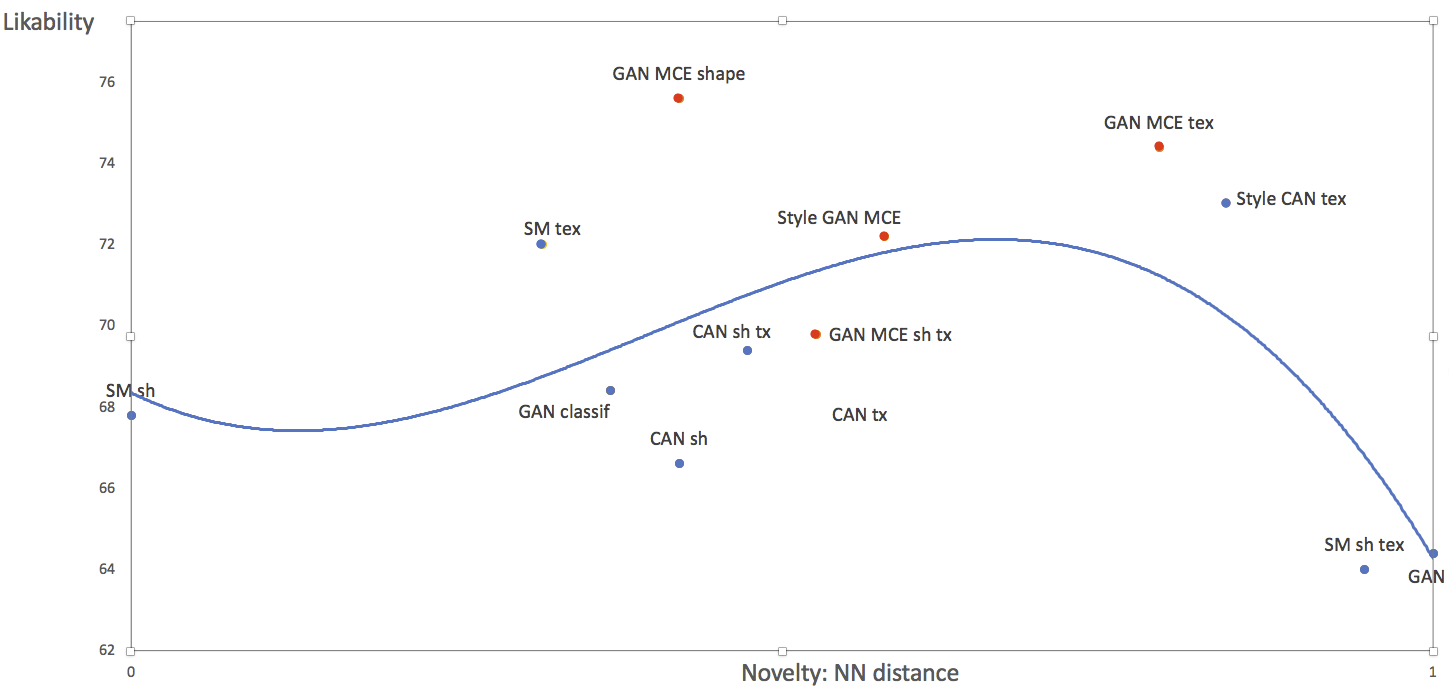}
  \end{center}
  \caption{Empirical approximation of the Wundt curve, drawing a relationship between novelty and appreciation. Models employing our proposed MCE losses appear in red, reaching a good compromise between novelty and likability. }
    \label{fig:Wundt}
\end{figure}

\begin{figure}[htb]
\begin{center}
\begin{tabular}{c}
\includegraphics[width=0.5\linewidth]{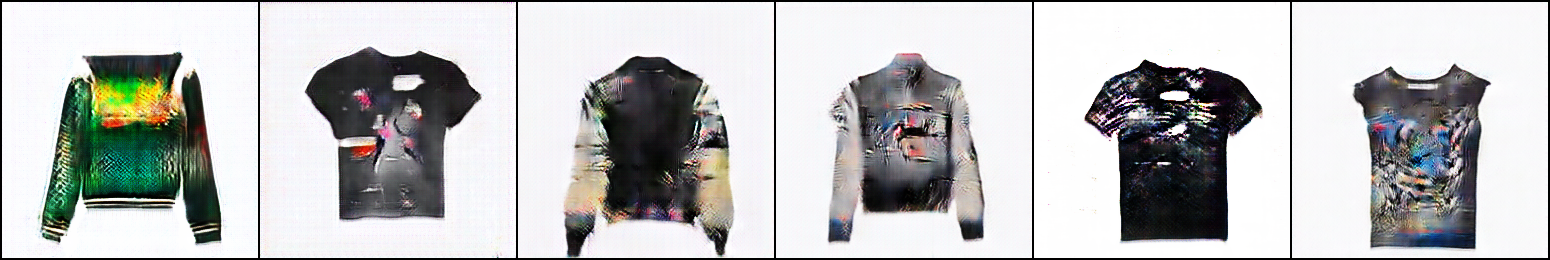}  \includegraphics[width=0.5\linewidth]{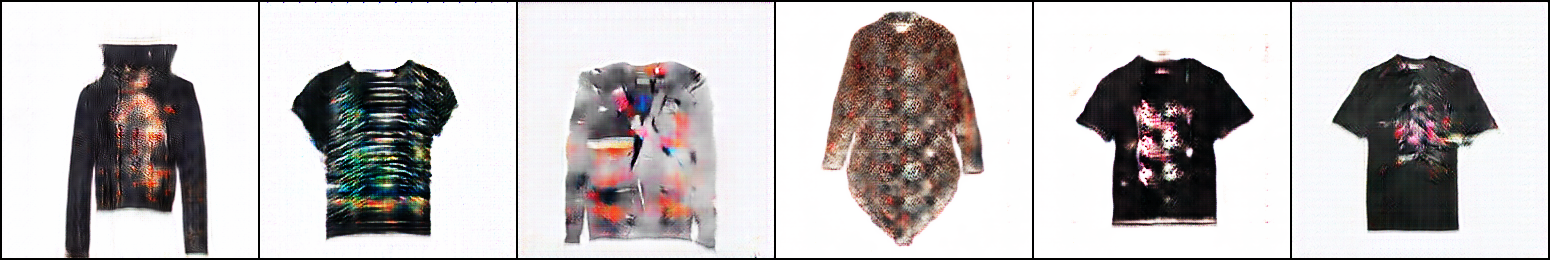}\\
\includegraphics[width=0.5\linewidth]{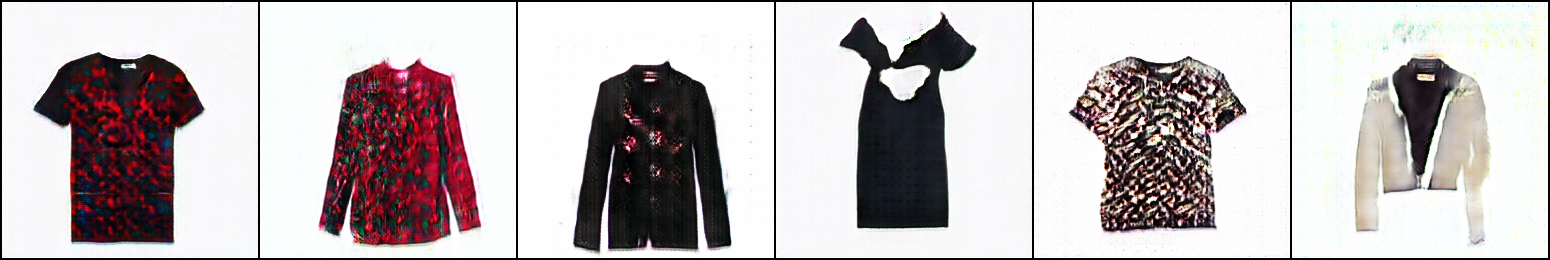}
\includegraphics[width=0.5\linewidth]{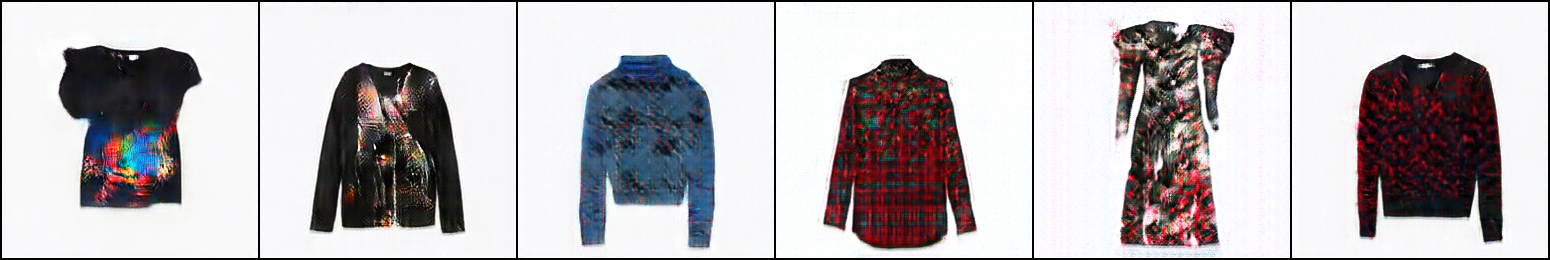}\\
\includegraphics[width=0.5\linewidth]{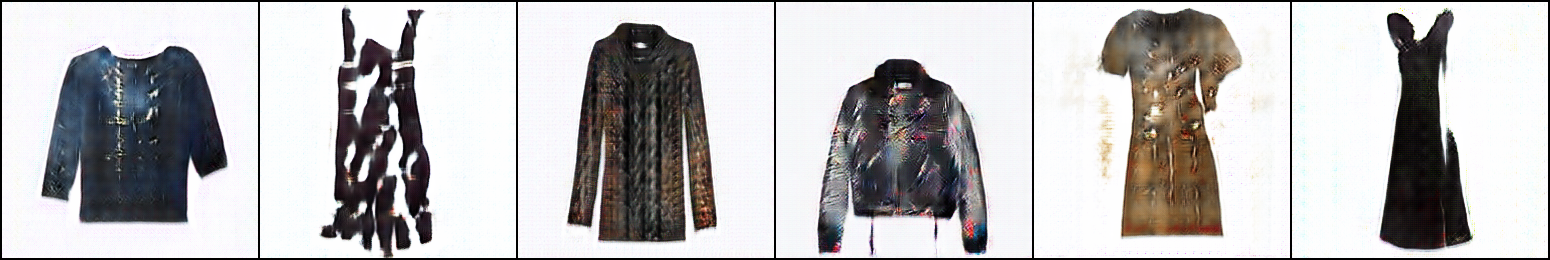}
\includegraphics[width=0.5\linewidth]{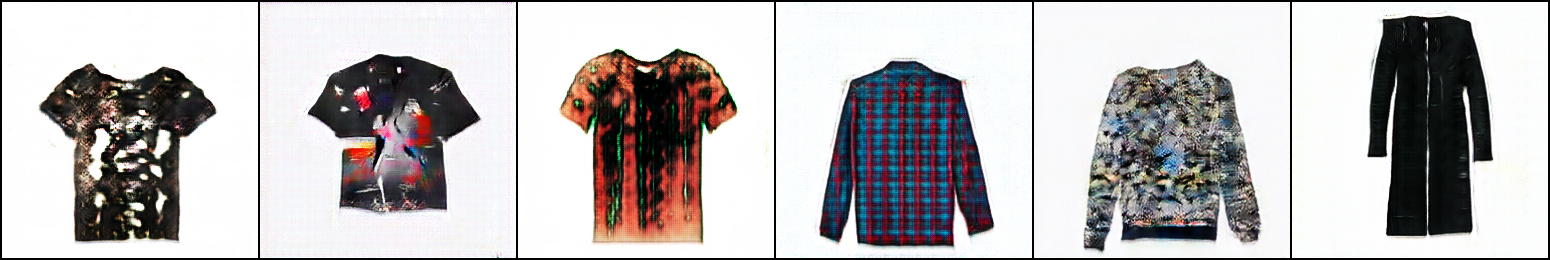}\\
\end{tabular}
\end{center}
\caption{Best generations of RTW items as rated by human annotators. Each question is in a row. Left: Q1: overall score, Q2: shape novelty, Q3: shape complexity, Right: Q4: texture novelty, Q5: texture complexity, Q6: Realism.}
\label{fig:best_human_per_qst}
\end{figure}

We may also recover the famous Wundt curve mentioned in the introduction by plotting the likability as a function of the novelty in Fig.~\ref{fig:Wundt}. \ccc{The novelty was difficult to assess by raters solely from generated images, therefore in this diagram, we give the novelty as the average distance from training images (NN metric).}
We note that the GAN with classification loss and models with Sharma-Mittal creativity exhibit low novelty. The most novel model is the classical GAN, however this kind of novelty is not very much appreciated by the raters. At the top of the curve, most of our new models based on the MCE \ccc{or SM} loss manage to find a nice compromise between novelty and hedonic value.

\begin{table}[htb]
\begin{center}

\begin{tabular}{cc}
\scriptsize{
\begin{tabular}{lcccccc}
\hline
Human & over- & shape & shape & tex. & tex. & real\\
Auto & all & nov. & comp. & nov. & comp. & fake\\
\hline
I sh score & 0.43 & \bf 0.65 & \bf 0.67 &0.33 &0.60 &0.30 \\
AM sh score & \bf 0.46 &\bf 0.66 &\bf 0.68 &0.38 &\bf 0.64 &0.28 \\
C sh score & \bf 0.48 &0.51 &0.55 &0.34 &0.50 &-0.06 \\
I tex score & \bf 0.48 &0.63 &0.62 &0.30 &0.55 &0.28 \\
AM score & 0.37 &0.47 &0.46 &0.16 &0.42 &0.22 \\
C tex score & \bf 0.48 &\bf 0.67 & \bf 0.71 & \bf 0.48 &\bf 0.64 &0.27 \\
N10 & \bf 0.46 &0.63 &0.65 &0.33 &0.58 &0.23 \\
Av. intensity & \bf 0.48 &\bf 0.65 & \bf 0.66 &0.35 &0.59 &0.23 \\
Skewness & 0.32 &0.62 & \bf 0.66 &0.32 &\bf 0.65 &0.31 \\
Darkness & 0.28 &0.54 &0.62 &0.32 &0.60 &\bf 0.43 \\
\hline
\end{tabular}
}&
\tiny{
\begin{tabular}{lcccccc}
\hline
Human & over- & shape & shape & tex. & tex. & real\\
Auto & all & nov. & comp. & nov. & comp. & fake\\
\hline
coat & 0.14 &0.37 &0.43 &0.21 &0.34 &0.27 \\
top & -0.15 &0.12 &0.16 &0.18 &0.07 &-0.08 \\
shirt & 0.43 &0.46 &0.37 &0.07 &0.30 &0.21 \\
jacket & 0.27 &0.36 &0.31 &0.08 &0.28 &-0.02 \\
t-shirt & 0.44 &0.49 &0.43 &0.31 &0.49 &0.08 \\
dress & 0.40 &0.49 &0.57 &0.36 &0.53 &0.12 \\
pullover & 0.24 &0.41 &0.50 &0.32 &0.47 &0.33 \\
dotted & 0.41 &0.35 &0.40 &0.26 &0.32 &0.25 \\
striped & 0.20 &0.16 &0.15 &0.08 &0.21 &0.11 \\
print & 0.42 &0.31 &0.26 &0.30 &0.35 &-0.04 \\
uniform & 0.31 &0.51 &0.54 &0.26 &0.51 &0.13 \\
tiled & 0.23 &0.17 &0.14 &-0.08 &-0.01 &0.22 \\
skin & -0.03 &0.32 &0.32 &0.30 &0.17 &0.17 \\
graphical & 0.34 &0.46 &0.47 &0.17 &0.46 &0.20 \\
\hline
\end{tabular}
}
\end{tabular}

\end{center}
\caption{Correlation scores between human evaluation ratings and automatic scores on the set of randomly sampled images of all models.}
\label{tb:corrAll}
\end{table}

\mypar{Correlations between human evaluation and automatic scores}

\begin{figure}[htb]
\begin{center}
\includegraphics[width=0.32\linewidth]{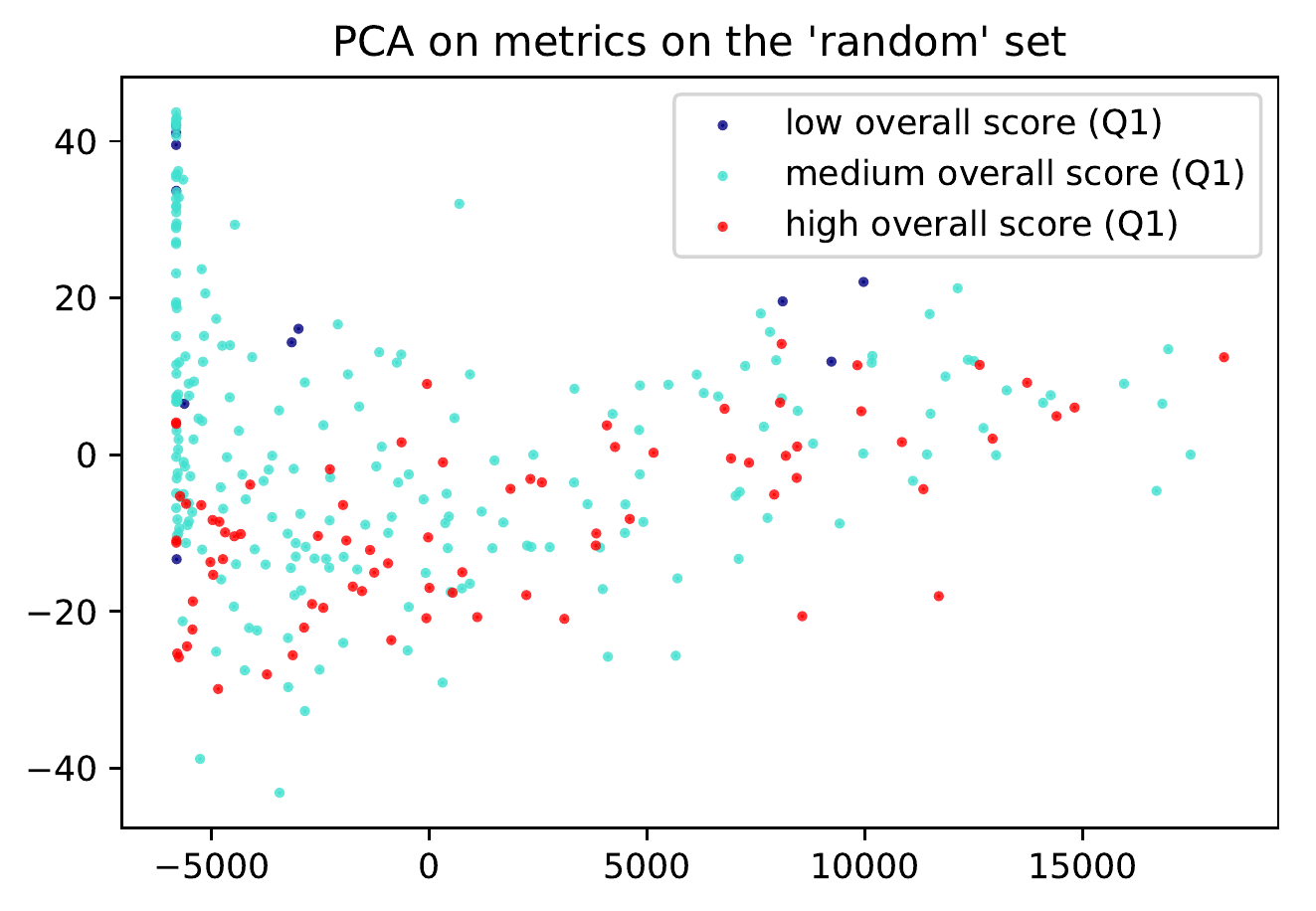}
\includegraphics[width=0.32\linewidth]{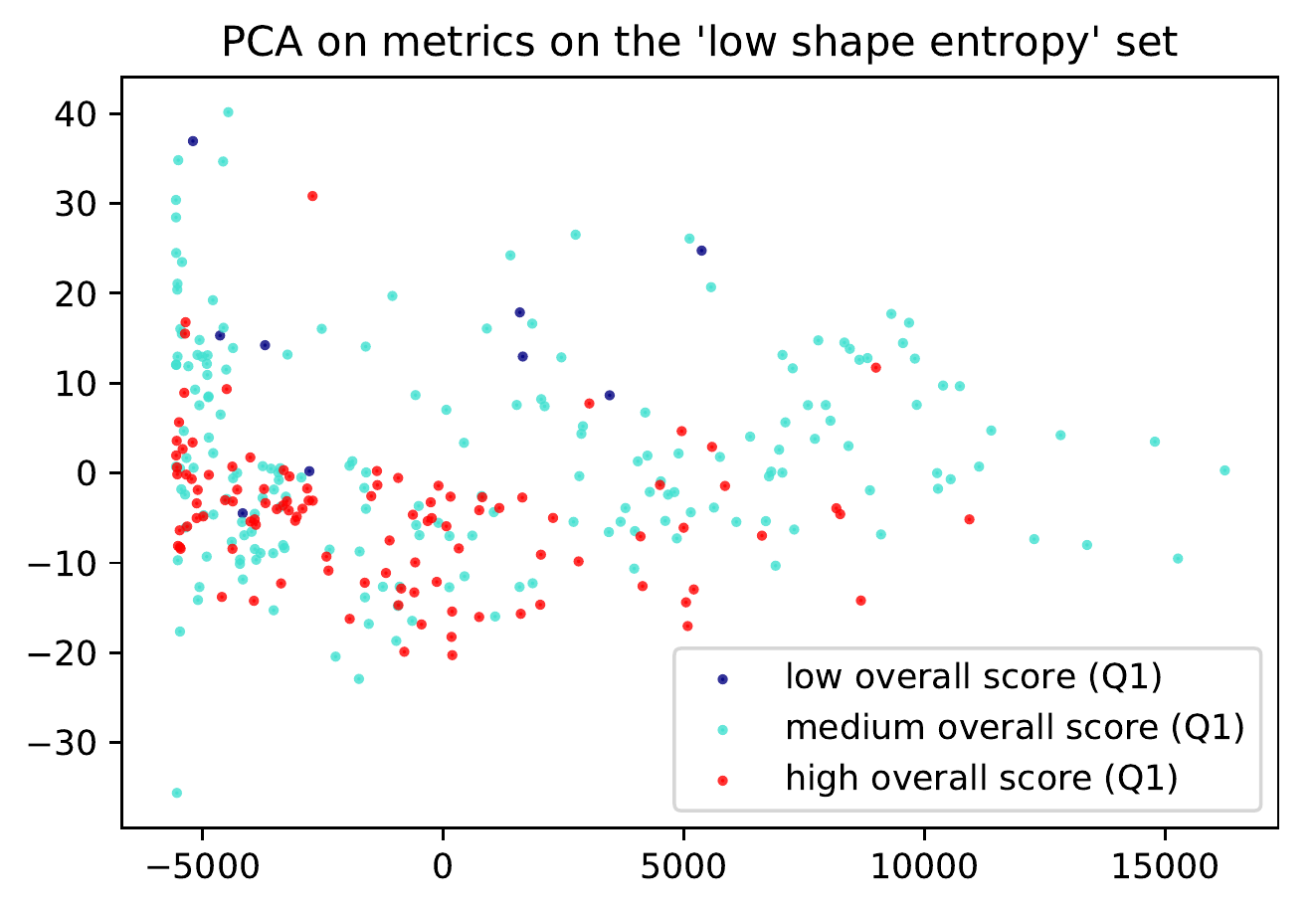}
\includegraphics[width=0.32\linewidth]{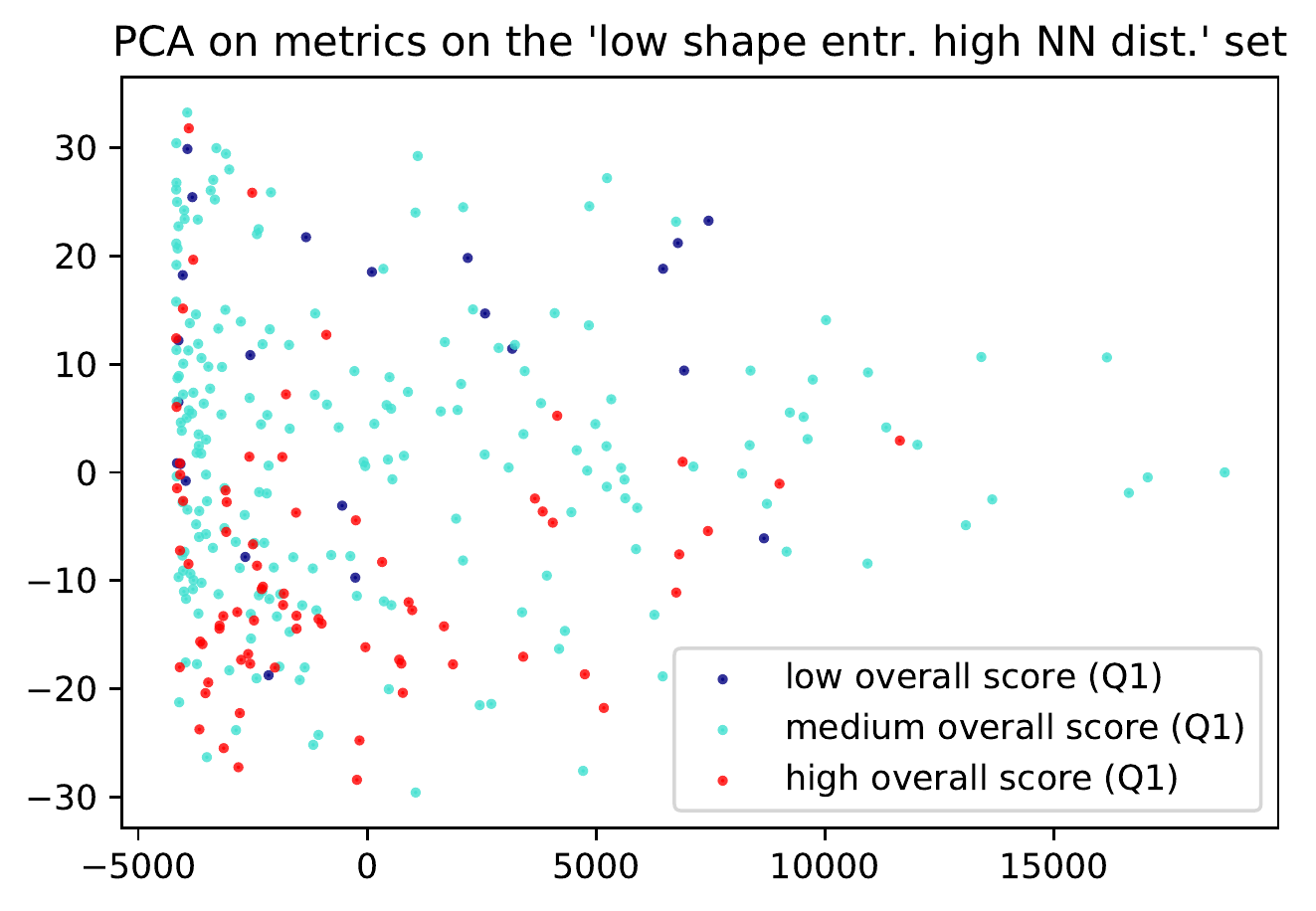}\\
\end{center}
\caption{PCA on image-wise automatic metrics (C tex, C shape, NN dist, Average
  intensity, Skewness, Darkness). Each metric was subtracted its mean value. The
  variables of the two first components are the Darkness and the Average
  intensity.}
\label{pca}
\end{figure}

From Table~\ref{tb:corrAll}, we see that the automatic metrics that correlate the most with the overall likability are \ccc{the classifiers confidence scores, the inception texture score, and the average intensity. The skewness measures correlates well with shape and texture complexity. The best correlated measure with real appearance is the Darkness as one could expect.
Figure \ref{pca} complements our analysis on correlations by presenting projections of the automatic metrics on generations from 3 sets (random, low shape entropy and low shape entropy and high NN distance). Each data point corresponding to one image is colored by a function of the human overall rating. The PCA reveals that the two principal components are mostly explained by the Darkness and Average intensity metrics. We observe from these diagrams that the different ratings are mixed but an overall pattern appears. }


\section{Conclusion}

We introduced a specific conditioning of convolutional Generative Adversarial Networks (GANs) on
texture and shape elements for generating fashion design images. While GANs with such classification loss offer realistic results, they tend to reconstruct the training images. Using creativity criteria, we learn to deviate from a reproduction of the training set. Our generalization of the existing CAN loss to the broader family of Sharma-Mittal divergence allows us to obtain more popular results.
We also propose a novel architecture -- our StyleGAN model -- conditioned on an input mask, enabling shape control while leaving free the creativity space on the inside of the item.
All these contributions lead to the best results according to our human evaluation study.
We manage to generate accurately $512\times512$ images, however we seek for better resolution, which is a fundamental aspect of image quality, in our future work. \ccc{Finally, there are still open research problems such as the automatic evaluation of generations and the improvement of the stability of generative adversarial networks. We plan to open source our Pytorch code upon paper acceptance.} 

\acks{
We would like to express our great appreciation to everyone involved in the RTW dataset creation. We are grateful to Alexandre Lebrun for his precious help for assessing the quality of our generations. We also would like to thank Julia Framel, Pauline Luc, Iasonas Kokkinos, Matthijs Douze, Henri Maitre, David Lopez Paz for interesting discussions. We finally acknowledge all participants to the evaluation study.}

\bibliography{clean_biblio}

\section{Appendix: Network architectures}
\label{subsec:networkArchs}
\subsection{Unconditional StackGAN}

\begin{figure*}[h]
\centering
\includegraphics[width=\textwidth]{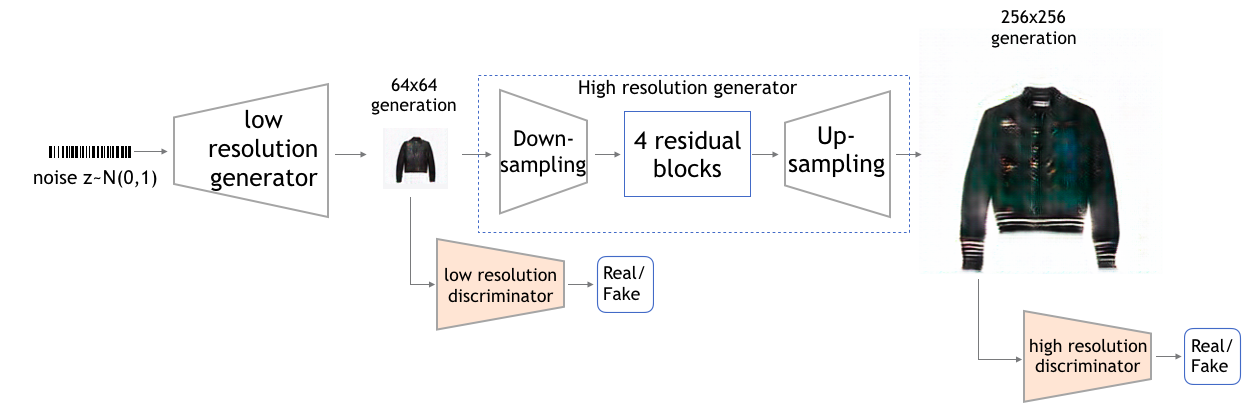}
\caption{StackGAN architecture adapted without text conditioning.}
\label{fig:stackGANarchitecture}
\end{figure*}

While the low resolution generator is a classical DCGAN architecture starting from noise only, the high resolution one takes an image as an input. We adapt its architecture as depicted in Fig.~\ref{fig:stackGANarchitecture} from the generator described in the fast style transfer model \citep{Johnson2016Perceptual} so that the output size is 256x256 given a 64x64 image. It uses convolutions for downsampling and transposed convolutions for upsampling, both with kernel size of 3, stride of 2, padding of 1, with no bias and followed by batchNorm (BN) then ReLU layers. The high resolution generator's architecture is described in Table ~\ref{tb:stackGANhighresgen}.

\begin{table}[htb]
\centering
\begin{tabular}{lccccc}
\hline
{Layer}   & \#output  & kernel & stride & padding   \\
{} &     channels & size &  &  \\ \hline
{conv}  & 64 & 7 & 1 & reflect(3)    \\
{conv}  & 128 & 3 & 2 & 1   \\
{conv}  & 128 & 3 & 2 & 1   \\
{Resnet block $\times 4$}  & 256 & - & - & -   \\
{convT}  & 256 & 3 & 2 & 1    \\
{convT} & 256 & 3 & 2 & 1    \\
{convT}  & 128 & 3 & 2 & 1    \\
{convT}  & 64 & 3 & 2 & 1  \\
{conv}  & 3 & 7 & 1 & reflect(3)   \\
\hline
\end{tabular}
\caption{Detailed architecture of the high resolution (second stage) generator of our proposed unconditional stackGAN architecture. Convolution layers (conv and convT) are followed by Batch normalization (BN) and ReLU, except for the last one, with no BN and Tanh non linearity.}
\label{tb:stackGANhighresgen}
\end{table}

\begin{table}[htb]
\centering
\begin{tabular}{lcccccc}
\hline
{} & layer &  \#input  & \# output  & ker. & str. & pad.   \\
{Branch} & &  &    & size & & \\ \hline
\multirow{3}{3em}{Shape mask branch}& conv & 3 & 64 & 5 & 2 & 2\\
 &conv & 64 & 128 & 5 & 2 & 2   \\
 &conv & 128 & 256 & 5 & 2 & 2  \\
\hline
\multirow{3}{3em}{Style noise branch}& fc & 100 & 1024 & 4 & 2 & 1 \\
 &convT & 64 & 64 & 4 & 2 & 1 \\
 &convT & 64 & 64 & 4 & 2 & 1 \\
 &convT & 64 & 64 & 4 & 2 & 1 \\ \hline
 {} & concat &  &  &  &  & \\ \hline
 &conv & 320 & 256 & 3 & 1 & 1   \\
 &conv & 256 & 512 & 3 & 2 & 1  \\
 &conv & 512 & 512 & 3 & 1 & 1  \\ \hline
 &convT & 512 & 256 & 4 & 2& 1   \\
 &convT & 256 & 128 & 4 & 2& 1   \\
  &convT & 128 & 128 & 4 & 2& 1  \\
  &convT & 128 & 64 & 4 & 2& 1   \\
  &convT & 64 & 3 & 5 & 1& 2    \\ \hline
\end{tabular}
\caption{Detailed architecture of our StyleGAN model. Convolution layers are followed by Batch normalization (BN) except for the last one. Conv layers are followed by leacky ReLU, ConvT layers by ReLU except the last layer where the non-linearity is a Tanh. }
\label{tb:styleGANhighresgen}
\end{table}

\subsection{StyleGAN architecture}

The architecture of the styleGAN model combines an input mask image with an input style noise as in the  style generator described in the structure style GAN architecture~\citep{Wang2016StyleStructureGAN}. This consists in upsampling the noise input with multiple transposed convolutions while downsampling the input mask with convolutions. Concatenating the two resulting tensors then performing a series of convolutional layers as shown in Table~\ref{tb:styleGANhighresgen} results in a $256\times256$ generation.

\begin{table*}
\centering
\begin{tabular}{lcccccc}
\hline
 & tex & over- & real & tex & shape & shape \\
 & nov. & all & appear & comp. & nov. & comp. \\ \hline
{tex nov.}  & 1.00   & 0.67   & -0.32   & 0.78   & 0.76   & 0.79  \\
{overall}  & 0.67   & 1.00   & -0.04   & 0.59   & 0.61   & 0.52  \\
{real}  & -0.32   & -0.04   & 1.00   & -0.47   & -0.25   & -0.48  \\
{tex comp.}  & 0.78   & 0.59   & -0.47   & 1.00   & \textbf{0.88}   & \textbf{0.93}  \\
{sh. nov}  & 0.76   & 0.61   & -0.25   & \textbf{0.88}   & 1.00   & \textbf{0.92}  \\
{sh. comp.}  & 0.79   & 0.52   & -0.48   & \textbf{0.93}   & \textbf{0.92}   & 1.00  \\
\hline
\end{tabular}
\caption{Auto-correlation between human experiments questions responses}
\end{table*}

\end{document}